\documentclass[10pt,twocolumn,letterpaper]{article}

\usepackage[pagenumbers]{wacv}
\definecolor{wacvblue}{rgb}{0.21,0.49,0.74}
\usepackage[pagebackref,breaklinks,colorlinks,allcolors=wacvblue]{hyperref}

\usepackage{colortbl}
\usepackage{pifont}
\newcommand{\cmark}{\textcolor{green}{\ding{51}}} 
\newcommand{\xmark}{\textcolor{red}{\ding{55}}}   
\definecolor{updates}{rgb}{0.1,0.75,0.24}

\title{Event-based Graph Representation with Spatial and Motion Vectors \\ for Asynchronous Object Detection}

\author{Aayush Atul Verma, Arpitsinh Vaghela, Bharatesh Chakravarthi, Kaustav Chanda, Yezhou Yang \\
Arizona State University \\
{\tt\small \{averma90, avaghel3, bshettah, kchanda3, yz.yang\}@asu.edu}\
}

\begin{document}
\twocolumn[{%
\renewcommand\twocolumn[1][]{#1}%
\maketitle
\begin{center}
    \centering
    \includegraphics[width=0.82\linewidth]{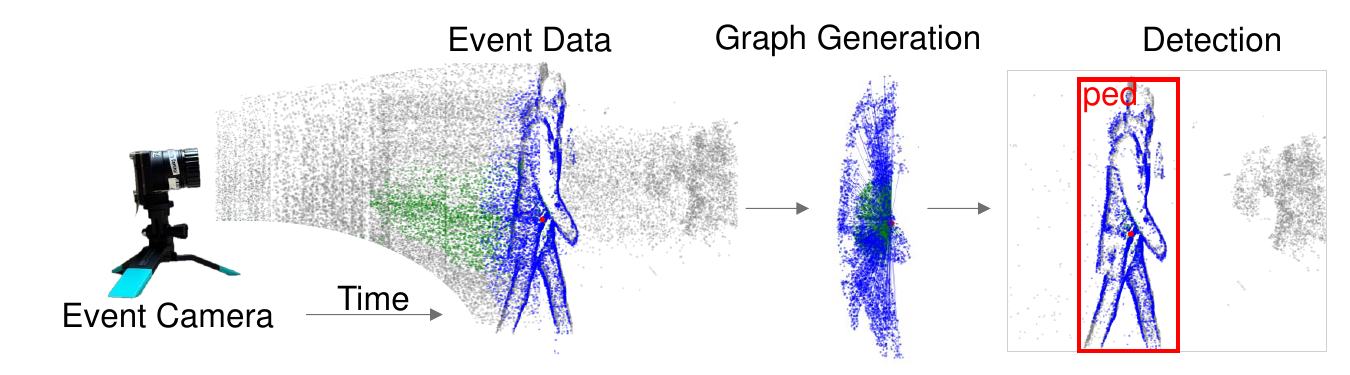}
    \captionof{figure}{\textbf{Overview of the Proposed Spatiotemporal Multi Graph Approach:} From raw event data (left), a spatiotemporal graph is constructed and processed using a novel graph learning strategy, enabling accurate detection (right) without dense conversion.}
    \label{fig_teaser}
\end{center}%
}]

\begin{abstract}
Event-based sensors offer high temporal resolution and low latency by generating sparse, asynchronous data. However, converting this irregular data into dense tensors for use in standard neural networks diminishes these inherent advantages, motivating research into graph representations. 
While such methods preserve sparsity and support asynchronous inference, their performance on downstream tasks remains limited due to suboptimal modeling of spatiotemporal dynamics.
In this work, we propose a novel spatiotemporal multigraph representation to better capture spatial structure and temporal changes. Our approach constructs two decoupled graphs: a spatial graph leveraging B-spline basis functions to model global structure, and a temporal graph utilizing motion vector-based attention for local dynamic changes. This design enables the use of efficient 2D kernels in place of computationally expensive 3D kernels. 
We evaluate our method on the Gen1 automotive and eTraM datasets for event-based object detection, achieving over a $6\%$ improvement in detection accuracy compared to previous graph-based works, with a $5\times$ speedup, reduced parameter count, and no increase in computational cost. These results highlight the effectiveness of structured graph modeling for asynchronous vision.
Project page: \href{https://eventbasedvision.github.io/eGSMV}{eventbasedvision.github.io/eGSMV}.
\end{abstract}    
\section{Introduction}
\label{sec:intro}
Event-based vision systems represent a paradigm shift in visual data capture~\cite{eventvisionsurvey, eventvisionsurvey2, chakravarthi2024recent}, offering high temporal resolution, sparse data, and asynchronous operation as an alternative to conventional frame-based imaging. Unlike traditional RGB cameras that capture frames at fixed intervals, event cameras record ``events'' only when pixel-level intensity changes occur, each providing visual information with microsecond-level latency~\cite{latency_async}. This event-driven approach results in a high dynamic range (up to 120 dB), low data rates, and reduced power consumption~\cite{low_power}. Importantly, it reduces motion blur and enhances responsiveness in challenging conditions, such as low light and high-speed motion~\cite{rebecq2019high}. These unique properties of sparsity and asynchrony make event cameras well-suited for applications in robotics~\cite{robotics_event, pedro}, autonomous driving~\cite{mvsec, dsec, sironi2018hats}, and surveillance~\cite{verma2024etram, aliminati2024sevd, dvsoutlab}, where rapid response to dynamic environments and low power consumption is essential.

Despite these advantages, most current research adapts frame-based architectures such as convolutional neural networks (CNNs)~\cite{1megapixel, inception_ssd, yolov3_events, astmnet} and vision transformers (ViTs)~\cite{rvt, ssm, sast, get-t}, originally designed for dense,  synchronous inputs. For compatibility, event streams must be converted into dense tensor representations. While this enables the use of powerful existing models, it comes at the cost of two core advantages of event data: its sparsity and asynchronous nature$!$ This densification increases computational overhead and limits the responsiveness and efficiency of the systems. These properties are especially critical in real-time and resource-constrained environments, where event-based vision holds great promise. These limitations highlight the need for alternative representations that can natively process the sparse, asynchronous data without sacrificing accuracy or efficiency.

To this end, approaches have focused on leveraging the inherent advantages of event data using spiking neural networks (SNNs)~\cite{snn1, snn2, snn3} and graph neural networks (GNNs)~\cite{aegnn}. SNNs are inspired by biological neurons and process information in an event-driven, asynchronous way, making them naturally compatible with neuromorphic hardware.
However, training deep SNNs to learn robust, generalizable patterns over extended timeframes remains challenging as discretization windows often misalign with the timing of event responses and result in temporal information loss. In contrast, GNNs offer a flexible alternative, with works like~\cite{aegnn, dagr} demonstrating the potential of representing raw event data as graphs in an event-by-event, asynchronous manner. This approach preserves temporal granularity and spatial sparsity while enabling adaptable spatiotemporal resolution, positioning GNNs as a promising direction for advancing event-driven applications.

AEGNN~\cite{aegnn} laid the foundation for implementing hierarchical learning in event graphs, demonstrating the potential of GNNs through efficient asynchronous update rules. Other works~\cite{graph3, graph4} show competitive performance in object recognition on short sequences but continue to struggle with more complex tasks. Performance especially falters in object detection and action recognition tasks, which require robust localization and handling long sequences. Recently,~\cite{dagr} achieved improved results by using deeper networks to enhance the capacity of these GNNs. However, this approach relies heavily on early temporal aggregation. This effectively compresses nodes into a single temporal voxel and disregards the granular temporal dynamics. These prior efforts demonstrate the potential of GNNs for event-based processing but reveal a performance ceiling, as none of these works fully capture the unique spatiotemporal dependencies intrinsic to event data.

In this work, we address these challenges by proposing a novel framework to model event-based GNNs while preserving their inherent sparse and asynchronous nature. To achieve this, we introduce a novel multigraph construction strategy to capture spatial and temporal relationships between nodes based on proximity as illustrated in \figureautorefname~\ref{fig_teaser}. 
We employ B-spline kernels to learn the spatial structure without requiring dense representations and incorporate attention-based motion vector learning for temporal modeling to capture long-sequence dependencies. 
Notably, our method achieves a \textbf{21\% improvement} in detection accuracy over AEGNN~\cite{aegnn} on the Gen1 dataset, along with a \textbf{13\% reduction} in computational complexity. On the eTraM dataset, we observe a \textbf{25\% improvement} in performance, demonstrating generalization across datasets. Compared to DAGr~\cite{dagr}, our approach achieves an \textbf{improvement of over 6\%} on Gen1 while maintaining similar computational requirements. Finally, we perform detailed ablations to evaluate each module, highlighting the effectiveness of our proposed spatiotemporal learning strategy.

Our key contributions can be summarized as follows:
\begin{enumerate}
    \item A novel multigraph representation, \textit{eGSMV}, is proposed to model raw event data that preserves its spatial sparsity and asynchronous high temporal resolution, thus enabling inference at a low-latency per event level.
    \item An efficient feature extraction approach is introduced, using anisotropic 2D kernels for spatial learning and motion-based attention for temporal learning, reducing computational demands per operation by up to $87.5$\%.
    \item We evaluate \textit{eGSMV} on the asynchronous event-based object detection task, demonstrating that it outperforms other asynchronous methods while requiring significantly lower computational cost and model size compared to synchronous methods.
\end{enumerate}

\section{Related Works}
Since the inception of event-based vision, various deep learning models have been increasingly explored to leverage unique characteristics. Initial approaches primarily relied on shallow learning techniques, including support vector machines~\cite{sironi2018hats} and filtering-based algorithms~\cite{HOTS, filter2, filter3} to extract relevant information from event streams. However, with the advancement of models like YOLO~\cite{yolo}, R-CNN~\cite{rcnn}, and RetinaNet~\cite{retinanet}, researchers began to investigate deep CNNs~\cite{Jiang2019MixedFF, conv2019async} to capitalize on their strengths.

\begin{figure*}[t]
    \centering
    \includegraphics[width=0.85\linewidth]{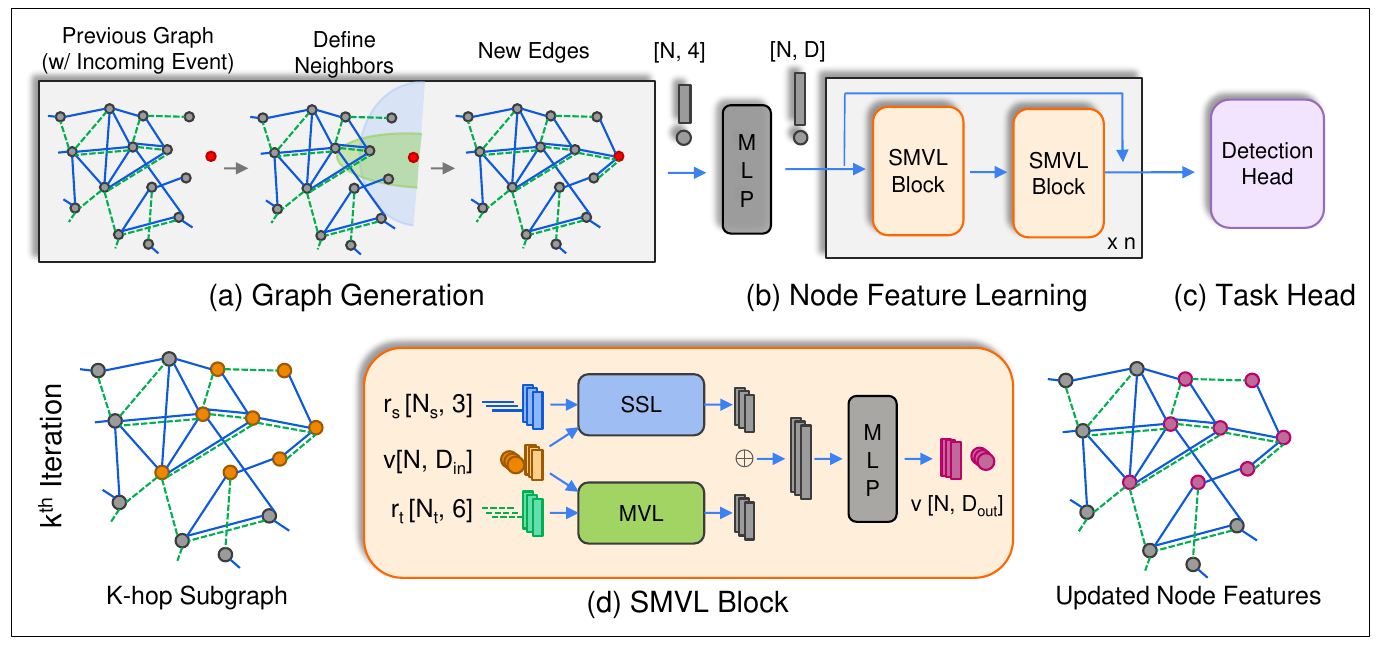}
    \caption{\textbf{Overview of \textit{eGSMV}:} (a) A new node asynchronously added to the graph by finding its spatial and temporal neighbors. (b)~Hierarchical update of node features to capture spatial and temporal relations through a series of SMVL blocks. (c) A specialized task head for event-driven object detection. (d) A single message passing step in the $k^{th}$ iteration to asynchronously update the k-hop subgraph.}
    \label{fig_methodology:architecture}
\end{figure*}

Since CNNs operate on independent frames, they disregard temporal dependencies in event data. To address this, models like RED~\cite{1megapixel} and ASTMNet~\cite{astmnet} introduced recurrent layers alongside CNNs to integrate temporal sequence modeling. Concurrently, transformer-based architectures~\cite{rvt, get-t, hmnet, peng2023better} showed promise in sequence modeling due to their capacity for capturing long-range dependencies. However, these methods require converting event data into dense tensor representations, sacrificing temporal resolution and sparsity. Furthermore, \cite{ssm} revealed that these RNN-based models show a drastic performance drop at frequencies different from their training frequency.

A complementary approach aims to retain the sparsity and asynchrony of event data through geometric-based methods and SNNs. While SNNs~\cite{snn1, snn2, snn3, snn4-eas-snn, snn5} support asynchronous processing on neuromorphic hardware, their lack of efficient learning rules limits scalability to complex tasks. Recently, geometric learning approaches attempt to address this gap by representing events as spatiotemporal point clouds~\cite{eventnet}, sparse submanifolds~\cite{eventscn}, or graphs~\cite{aegnn, dagr, graph1, graph2, graph3, bi2019graph, graph4, graph5} and processing them with specialized neural networks. Graph neural networks (GNNs) have shown promise, achieving strong performance in object recognition~\cite{graph3, bi2019graph, graph4}, object detection~\cite{dagr, aegnn}, and motion segmentation~\cite{graph_segment, motionseg_graphcut} while preserving data sparsity.~\cite{aegnn} demonstrates that GNNs trained synchronously can run asynchronously during inference to achieve the same results. Further improving upon it,~\cite{dagr} introduced optimized look-up table-based message-passing techniques to enable deep, high-capacity networks. 
However, these optimizations rely on early temporal aggregation, which sacrifices granularity and limits their suitability in tasks requiring a fine-grained temporal understanding. 
Furthermore, none of these approaches explicitly models the spatial and temporal characteristics unique to event data, which has a sparse spatial distribution and high temporal density. 
\cite{edgcn} aims to learn such attributes for each vertex to achieve a better representation. However, they rely on full graph reconstruction in every layer and voxelize events within $>25ms$ time windows. This introduces latency and makes them particularly unsuitable for asynchronous processing. To address these limitations, our work presents a novel multigraph framework to capture the dependencies effectively without additional computational costs.
\section{Spatial and Motion Vector Graph Learning}
We present \textbf{\textit{eGSMV}}, a novel three-stage architecture that models events as a spatiotemporal multigraph. As illustrated in \figureautorefname~\ref{fig_methodology:architecture}~(a-c), it consists of the following stages: 
\((i)\)~\textbf{Graph Representation} – a unique graph construction strategy to represent event data in 3D spatiotemporal space. (Section~\ref{graph_representation}) \((ii)\)~\textbf{Node~Feature Learning} – a series of message-passing steps to hierarchically capture both the spatial structure and the motion vector of each node. (Section~\ref{feature_learning}) \((iii)\)~\textbf{Downstream Task Head} – a specialized task head designed for object detection. (Section~\ref{downstream_task_head}) 

\subsection{Graph Representation of Event Data}
\label{graph_representation}
Event cameras consist of independent pixels that trigger events whenever they detect a change in brightness. Each event encodes the pixel position \((x_i, y_i)\), timestamp \(t_i\) with microsecond-level resolution, and polarity \(p_i \in \{-1, 1\}\), indicating the direction of change. An event stream within a time window \(\Delta T\) can therefore be represented as an ordered list of tuples,
\begin{equation}
\{e_{i}\}^{N}_{i=1} = {(x_{i}, y_{i}, t_{i}, p_{i})^{N}_{i=1}}
\label{eq:event_representation}
\end{equation}

\begin{figure}[]
  \centering
   \includegraphics[width=0.9\linewidth]{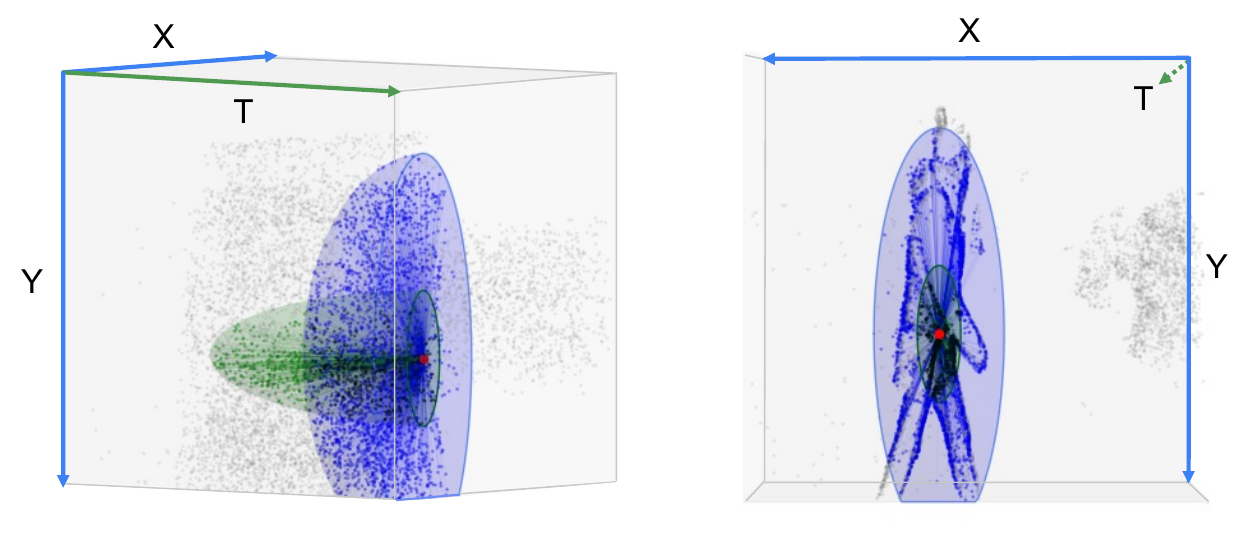}
   \caption{\textbf{Graph Representation:} Different views of the neighbor selection strategy in spatiotemporal space. (Green denotes temporal neighbors while blue denotes spatial neighbors of the red event)}
   \label{fig_methodoloy:neighbor_selection}
\end{figure}

This encoding makes raw event data spatially sparse yet temporally dense due to its high temporal resolution. Combined with the unique characteristic of encoding different information across the spatial and temporal dimensions, this data structure requires a precise method to capture both relationships effectively. To address this need, we introduce a novel strategy to identify relevant neighbors for each event, as depicted in Figure~\ref{fig_methodoloy:neighbor_selection}. 

Our event multigraph \(\mathcal{G} = \{\mathcal{V}, \mathcal{E}_{s}, \mathcal{R}_{s}, \mathcal{E}_{t}, \mathcal{R}_{t}\} \), represents each event \( e_i \) as a node \( v_i \in \mathcal{V}\) positioned at \((x_i, y_i, t_i)\) in the \(\mathbb{R}^3\) spatiotemporal space. A directed edge \( e_{s_{ij}} = (i, j) \in  \mathcal{E}_{s} \) indicates that $v_j$ is a spatial neighbor of $v_i$ with attribute \( r_{s_{ij}} \in \mathcal{R}_{s} \). Similarly a directed edge \( e_{t_{ij}} = (i, j) \in  \mathcal{E}_{t} \) represents a temporal relationship, where $v_j$ is a temporal neighbor of $v_i$ with attribute \( r_{t_{ij}} \in  \mathcal{R}_{t} \). All edges in the multigraph are directed, with each edge $e_{ij}$ signifying a directional relationship in which $v_j$ is a neighbor of $v_i$.

To capture the local spatial structure, we define \textit{spatial neighbors} within an ellipsoidal vicinity, with the semi-major axis on the XY spatial plane and the semi-minor axis along the temporal dimension. This approach ensures that each node aggregates spatial information to enhance its understanding of the local structure. A spatial edge $(i, j) \in \mathcal{E}_{s}$ is constructed if,
\begin{equation}
\frac{\| \mathbf{v}_i^{xy} - \mathbf{v}_j^{xy} \|}{R_{XY}} + \frac{|t_i - t_j|}{R_t} < 1 \quad \text{and} \quad t_i < t_j,
\label{eq:event_representation}
\end{equation}
where $\|.\|$ denotes Euclidean distance and $\mathbf{v}_i^{xy}$ represents the $XY$ coordinate of node $v_i$. For spatial neighbors, $R_{XY}$ is set to $4\%$ of the input dimension, and $R_t$ represents a radius of 5$ms$.

To account for motion changes over time, we define a separate set of neighbors, referred to as \textit{temporal neighbors}, to capture temporal dependencies. Here, we employ the ellipsoidal strategy with an inverted orientation: the semi-major axis lies along the temporal dimension, with $R_t$ set to 40$ms$, and the semi-minor axis lies on the $XY$ spatial plane, with $R_{XY}$ set to 1\% of the input dimension. We limit each node to $16$ spatial and $12$ temporal neighbors to ensure computational efficiency and avoid overfitting while preserving relevant information. Finally, the initial node features are defined as, $\mathbf{x_i} = (x_i, y_i, t_i, p)$, normalized to the range $\{-1, +1\}$ for each node $v_i$.
Defining spatial and temporal neighbors enables \textit{eGSMV} to effectively capture complex spatiotemporal dependencies within event data, laying a strong foundation for downstream tasks. The values for $R_{XY}$, $R_{t}$, and the number of neighbors are chosen based on hyperparameter optimization to balance computational efficiency, avoid overfitting, and maximize performance.

\subsection{Node Feature Learning}
\label{feature_learning}
In this stage, \textit{eGSMV} hierarchically learns feature representations by aggregating information from its neighbors, ensuring each node is contextually rich for downstream tasks. First, the initial node features $\mathbf{x_i}$ are projected to a higher-dimensional space using a multi-layer perceptron (MLP). This richer representation is then passed through a series of spatial and motion vector learning blocks (SMVL). As shown in \figureautorefname~\ref{fig_methodology:architecture}~(d), SMVL independently models node's spatial and temporal neighborhoods independently. It consists of two components: spatial structure learning (SSL) to capture local spatial hierarchies using the spatial neighborhood and motion vector learning (MVL) to capture temporal dependencies using the temporal neighborhood. 
Each of these components updates node representations to capture the specific context they are designed to model. This is done by aggregating message vectors from the node's spatial or temporal neighbors and corresponding edge features.
Formally, for each node $i$, an aggregated message vector \( \mathbf{m}_i^{(n)} \) at the $n^{th}$ step is obtained as:
\begin{equation}
\mathbf{m}_i^{(n)}=\text{AGGREGATE}^{(n)} \left( \left\{ \mathbf{v}_j^{(n-1)}, r_{ij}:j \in \mathcal{N}(i) \right\} \right),
\label{eq:event_representation}
\end{equation}
where \( \mathbf{v}_j^{(n-1)} \) represents the features of neighboring nodes and \(r_{ij}\) denotes any edge features, such as spatial or temporal weights. This aggregated message is then used to update the node's feature vector \( \mathbf{v}_i^{(n)} \) via:
\begin{equation}
\mathbf{v}_i^{(n)} = \text{UPDATE}^{(n)} \left( \mathbf{v}_i^{(n-1)}, \mathbf{m}_i^{(n)} \right)
\label{eq:event_representation}
\end{equation}
Since spatial and temporal neighbors are processed separately, we get two updated node features capturing the spatial and temporal contexts. These features are then fused at the node level to learn a feature with a richer spatiotemporal context. By iterating over multiple message-passing steps, each node learns to hierarchically incorporate increasingly complex spatiotemporal information from its neighborhood to ensure contextually rich features well-suited for downstream tasks.
In the following subsections, the SSL, MVL, and feature fusion components are described in detail.
\subsubsection{Spatial Structural Learning}
The primary objective of the SSL block is for every node to learn about its local spatial structural properties in the 2D spatial plane. An \textit{Anisotropic Spline Convolution} kernel with dimension ($k\times k \times 1$) is utilized for this purpose, which operates across the $XY$ spatial plane with a depth of 1 in the temporal dimension. This allows the kernel to capture spatial dependencies without extending across time, effectively making it a 2D operation. Compared to isotropic 3D kernels used in prior methods, such an anisotropic design reduces computational complexity by up to $87.5\%$ (for $k=8$) with a lower parameter count, making it efficient for learning spatial hierarchies. 

Each node’s spatial structure is encoded through edges linking it to spatial neighbors, with edge features defined by the normalized cartesian difference in position between nodes $i$ and $j$:
\begin{equation}
\mathbf{e}_{ij}^{\text{spatial}} = (\Delta x, \Delta y, \Delta t), \quad 
\label{eq:event_representation}
\end{equation}
where $\Delta x = x_j - x_i, \; \Delta y = y_j - y_i, \; \Delta t = t_j - t_i$.
This spatial encoding enables the kernel to aggregate features relative to each node’s local structure, enabling fine-grained spatial dependency learning. Multiple rounds of message passing, allows each node to learn from a larger subgraph, giving it a deeper understanding of the global structure. By explicitly focusing on the spatial plane, SSL aligns with the inductive bias in CNNs, allowing it to achieve similar benefits in learning the local structure while operating in a sparse domain.

\subsubsection{Motion Vector Learning} The MVL block captures motion patterns by aggregating information from temporal neighbors - nodes close in space but spanning previous time steps. This structure allows each node to develop a coarse understanding of how motion has evolved, including changes in position, velocity, and brightness. To achieve this, edge features represent positional and dynamic information between a node and its temporal neighbors:
\begin{equation}
\mathbf{e}_{ij}^{temporal} = {(\Delta x, \Delta y, \Delta t, \dfrac{\Delta x}{\Delta t}, \dfrac{\Delta y}{\Delta t}, \Delta p)},
\label{eq:event_representation}
\end{equation}
Here, the terms $(\Delta x, \Delta y)$ provide spatial displacements that act as relative positional embeddings, while $(\Delta x/\Delta t, \Delta y/\Delta t)$ characterize motion as velocity components in the $X$ and $Y$ directions. Additionally, $\Delta p$ represents the change in polarity, capturing potential relative motion through the variation in the scene’s brightness or contrast. Together, these features create a comprehensive temporal profile of each node’s motion context.

A coherent understanding of motion over sequential data requires each node to selectively attend to its temporal neighbors based on their relevance. By applying multi-head attention, we dynamically weigh each neighbor's contribution according to its prior significance in the motion trajectory. This approach allows nodes to enrich their temporal understanding and enhance the network’s ability to interpret complex motion patterns across asynchronous event streams. This attention is implemented with  GATv2~\cite{gatv2} to adaptively focus on key temporal features.

\subsubsection{Feature Fusion} 
To maintain low computational overhead for efficient end-to-end asynchronous processing, the spatial features from the SSL block and temporal features from the MVL block are fused via a simple concatenation followed by a MLP. 
This deliberately lightweight design choice preserves full spatial and temporal information while avoiding the extra overhead introduced by cross-attention or other Transformer-based fusion methods, providing an efficient yet comprehensive spatiotemporal representation.

\subsection{Downstream Task Head}
\begin{figure}[]
  \centering
   \includegraphics[width=0.9\linewidth]{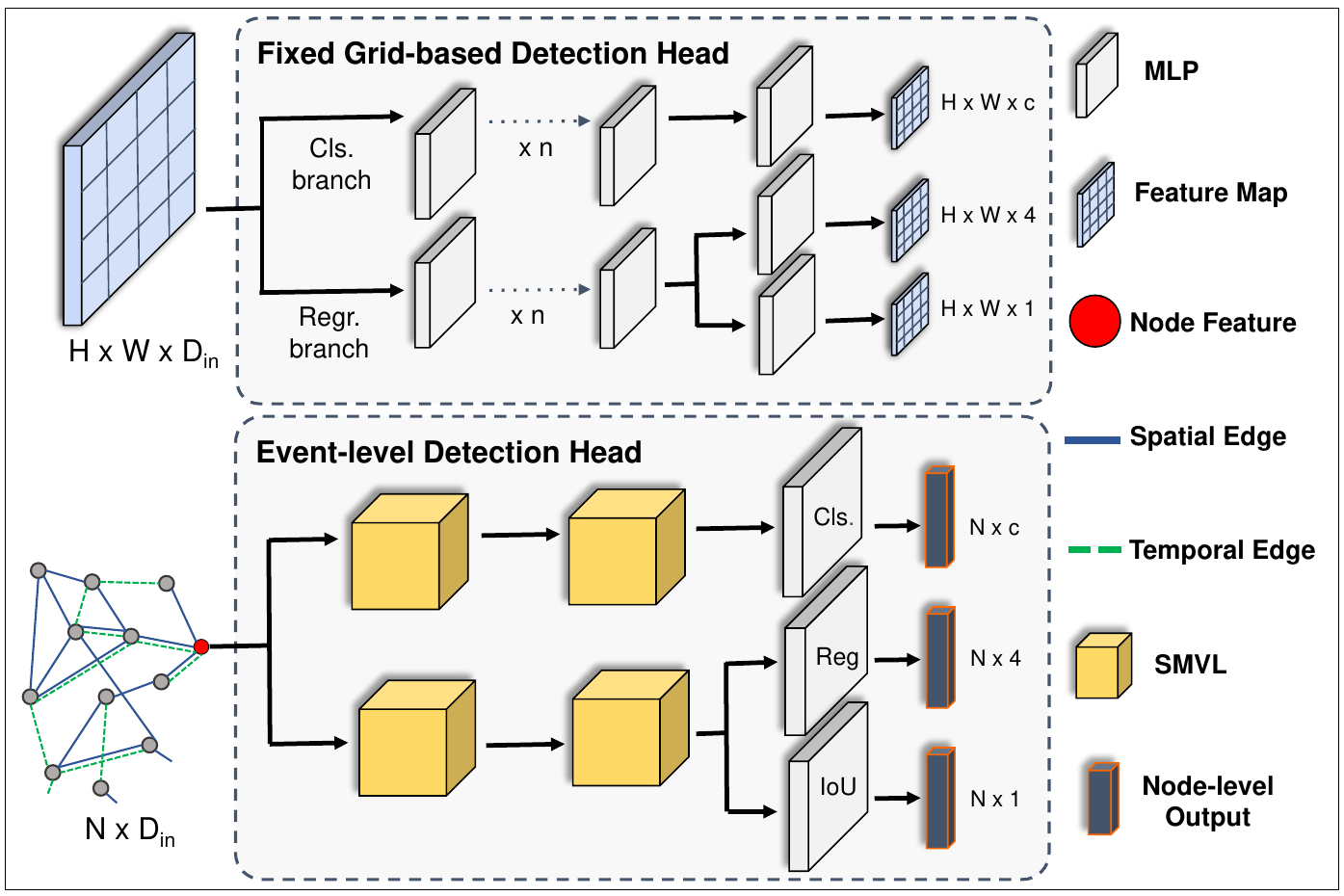}
   \caption{\textbf{Event-level Detection Head:} Comparing our async. event-driven detection head vs. fixed grid-based detection head.}
   \label{fig_methodoloy:detection_head}
\end{figure}

\begin{table*}[t]
\centering
\resizebox{0.95\textwidth}{!}{%
\begin{tabular}{lcccccccc}
\toprule \hline
\multicolumn{1}{c}{\textbf{}} &
  \textbf{} &
  \multicolumn{1}{l}{\textbf{}} &
  \textbf{} &
  \multicolumn{2}{c}{\textbf{Gen1}} &
  \multicolumn{2}{c}{\textbf{eTraM}} &
  \textbf{} \\ \hline
\multicolumn{1}{c|}{\textbf{Methods}} &
  \textbf{Representation} &
  \textbf{Backbone} &
  \multicolumn{1}{c|}{\textbf{Async.}} &
  \textbf{mAP $\uparrow$} &
  \multicolumn{1}{c|}{\textbf{MFLOPs/ev $\downarrow$}} &
  \textbf{mAP@50 $\uparrow$} &
  \multicolumn{1}{c|}{\textbf{MFLOPs/ev $\downarrow$}} &
  \textbf{Params (M) $\downarrow$} \\ \hline \midrule
\multicolumn{1}{l|}{RED~\cite{1megapixel}} &
  Event Volume &
  CNN+RNN &
  \multicolumn{1}{c|}{\xmark} &
  0.400 &
  \multicolumn{1}{c|}{4712} &
  \cellcolor{red!8}0.491 &
  \multicolumn{1}{c|}{$>$ 10000} &
  24.1 \\
\multicolumn{1}{l|}{ASTMNet~\cite{astmnet}} &
  Event Volume &
  CNN+RNN &
  \multicolumn{1}{c|}{\xmark} &
  0.467 &
  \multicolumn{1}{c|}{2930} &
  - &
  \multicolumn{1}{c|}{ - } &
  $>$ 100* \\
\multicolumn{1}{l|}{YOLOv3\_DVS~\cite{yolov3_events}} &
  Event-Histogram &
  CNN &
  \multicolumn{1}{c|}{\xmark} &
  0.312 &
  \multicolumn{1}{c|}{11100} &
  0.178 &
  \multicolumn{1}{c|}{$>$ 10000} &
  63.7* \\
\multicolumn{1}{l|}{RVT~\cite{rvt}} &
  Event-Histogram &
  Transformer+RNN &
  \multicolumn{1}{c|}{\xmark} &
  0.472 &
  \multicolumn{1}{c|}{3520*} &
  \cellcolor{red!20}0.539 &
  \multicolumn{1}{c|}{$>$ 10000} &
  18.5 \\
\multicolumn{1}{l|}{SSM~\cite{ssm}} &
  Event-Histogram &
  Transformer+SSM &
  \multicolumn{1}{c|}{\xmark} &
  \cellcolor{red!8}0.477 &
  \multicolumn{1}{c|}{3520*} &
  - &
  \multicolumn{1}{c|}{-} &
  18.2 \\
\multicolumn{1}{l|}{SAST~\cite{sast}} &
  Event-Histogram &
  Transformer+RNN &
  \multicolumn{1}{c|}{\xmark} & \cellcolor{red!20}0.482
   & 
  \multicolumn{1}{c|}{2400}  & 
  - &
  \multicolumn{1}{c|}{-} &
  18.9 \\ \midrule
\multicolumn{1}{l|}{Asynet~\cite{asynet}} &
  Event-Histogram &
  CNN &
  \multicolumn{1}{c|}{\cmark} &
  0.129 &
  \multicolumn{1}{c|}{205} &
  - &
  \multicolumn{1}{c|}{-} &
  11.4 \\
\multicolumn{1}{l|}{EAS-SNN~\cite{snn4-eas-snn}} &
  ARSNN &
  SNN &
  \multicolumn{1}{c|}{\cmark} &
  \cellcolor{blue!24}0.375 &
  \multicolumn{1}{c|}{-} &
  - &
  \multicolumn{1}{c|}{-} &
  25.3 \\
\multicolumn{1}{l|}{VC-DenseNet~\cite{vc-densenet}} &
  Voxel Cube &
  SNN &
  \multicolumn{1}{c|}{\cmark} &
  0.189 &
  \multicolumn{1}{c|}{-} &
  - &
  \multicolumn{1}{c|}{-} &
  8.2 \\ \midrule
\multicolumn{1}{l|}{NVS-S~\cite{nvs}} &
  Graph &
  GNN &
  \multicolumn{1}{c|}{\cmark} &
  0.086 &
  \multicolumn{1}{c|}{7.80} &
  - &
  \multicolumn{1}{c|}{-} &
  \cellcolor{yellow!40}0.9 \\
\multicolumn{1}{l|}{AEGNN~\cite{aegnn}} &
  Graph &
  GNN &
  \multicolumn{1}{c|}{\cmark} &
  0.163 &
  \multicolumn{1}{c|}{\cellcolor{orange!8}5.26} &
  0.180 &
  \multicolumn{1}{c|}{29.8} &
  20.1 \\
\multicolumn{1}{l|}{DAGr~\cite{dagr}} &
  Graph &
  GNN &
  \multicolumn{1}{c|}{\cmark} &
  \cellcolor{teal!8}0.212/0.304 &
    \multicolumn{1}{c|}{6.27/4.58} &
  \multicolumn{1}{c}{-} &
  \multicolumn{1}{c|}{-} &
  34.6\textasciicircum \\ \midrule
\multicolumn{1}{l|}{\textbf{eGSMV (ours)}} &
  Graph &
  GNN &
  \multicolumn{1}{c|}{\cmark} &
  \cellcolor{teal!20}0.371 &
  \multicolumn{1}{c|}{ \cellcolor{orange!20}4.5} &
  \cellcolor{teal!20}0.431 &
  \multicolumn{1}{c|}{\cellcolor{orange!20}26.1} &
  \cellcolor{yellow!10}5.6 \\
  \hline \bottomrule 
\end{tabular}%
}
\caption{\textbf{Comparisons with State-of-the-Art Methods:} We report mAP for Gen1 and mAP with an IoU threshold of 50\% for eTraM. Performance against dense representations and asynchronous methods are presented. (*) suggests that these values were not directly available and were estimated based on other sources. (\textasciicircum) suggests that the values are not representative for the above comparison.}
\label{tab:baseline_table}
\end{table*}

\textbf{Event-level Sparse Object Detection}.
\label{downstream_task_head}
Our approach is designed to leverage event data's sparse asynchronous nature and eliminate the need to convert streams into a dense feature map or a fixed grid representation, as seen in previous works
Figure~\ref{fig_methodoloy:detection_head}. The head maintains the point-wise representation to process the sequence in its original structure such that each event predicts its associated object class and bounding box.
Similar to existing architectures~\cite{yolox2021}, the detection head consists of a classification and a regression branch, with the former predicting a \(n+1\) multiclass probability distribution \(\{p_{1},.., p_{n+1}\}\), where \(n+1\) accounts for the number of object classes and a background class. The regression branch, in turn outputs normalized bounding boxes \(\{x', y', w', h'\}\) and an object IoU score where,
    \begin{equation}
         x' = \dfrac{x_{gt} - x_{\text{pos}}}{w_{0}}  \quad
         y' = \dfrac{y_{gt} - y_{\text{pos}}}{h_{0}} 
    \label{eq:event_representation}
    \end{equation}
    \begin{equation}
         w' = log (\dfrac{w_{gt}}{w_{0}}) \quad
         h' = log (\dfrac{h_{gt}}{h_{0}})
    \label{eq:event_representation}
    \end{equation}
\(\{x_{gt}, y_{gt}, w_{gt}, h_{gt}\}\) represents the ground truth center, width, height, while \(\{w_{0}, h_{0}\}\) are normalized scale factors.

Since spatially and temporally proximate events correspond to the same object, redundant detections occur. To address this, we introduce active regions formed by pooling these proximal events, followed by non-maximum suppression. This allows \textit{eGSMV} to perform asynchronous and sparse operations at the inference step as well, unlike prior works that rely on dense layers.
This enables the framework to make efficient end-to-end predictions in its native sparse format without dense transformations at any step.

\section{Experiments}
\label{section:experiments}
In this section, we present evaluations and ablations of \textit{eGSMV} on real-world Gen1 Automotive Detection~\cite{gen1} and eTraM~\cite{verma2024etram} datasets. To highlight its effectiveness, we compare it with existing frequency-based synchronous and graph-based asynchronous methods, examining model complexity in terms of the number of floating-point operations (FLOPs) and trainable parameters. 

\subsection{Experimental Setup}
\textbf{Datasets.} The Gen1 and eTraM datasets present distinct challenges due to their different perspectives and characteristics. This allows us to assess the performance in differing graph structures. Gen1 captures data from an ego-motion perspective and comprises $39$ hours of events. It has a $304\times240\,\text{px}$ resolution and contains $2$ object classes. The labeling frequency for this dataset is 20$Hz$. On the contrary, eTraM has $10$ hours from a static perspective, leading to a minimal amount of background events and more sparse data. It has a $1280\times720\,\text{px}$ resolution and includes $8$ labeled object classes. The dataset has a labeling frequency of 30$Hz$ and contains $2M$ bounding boxes.

\textbf{Implementation Details.}  
All graph networks in this work are implemented using the PyG library~\cite{pyg} and trained with the lightning framework~\cite{lightning}. We use the AdamW optimizer with a learning rate of $4\times 10^{-4}$ and a weight decay of $10^{-4}$. A OneCycleLR learning rate schedule is applied over $175k$ steps with linear annealing, updating at each training step. Training is conducted with a batch size of $24$ for both datasets, and each constructed event sequence represents a $100ms$ time window. For data augmentation, random translation and cropping are applied during training, as detailed in the supplementary material. 
Training with mixed precision on an Nvidia H100 GPU requires approximately $2$ days for both datasets.

\textbf{Metrics.} In line with prior studies,~\textit{eGSMV} is evaluated on mean average precision (mAP)~\cite{coco} for Gen1 and mAP with an IoU threshold of $50\%$ for eTraM (\tableautorefname~\ref{tab:baseline_table}). The average MFLOPs for adding a new event to an existing graph and the parameter count of the model are reported.

\subsection{Comparison with SOTA}

In this section, we compare our proposed multigraph framework, \textit{eGSMV}, against state-of-the-art methods on both the Gen1~\cite{gen1} and eTraM~\cite{verma2024etram} dataset as summarized in \tableautorefname~\ref{tab:baseline_table}. We choose the model that performs best on the validation set and present results from the test set.

\begin{figure}[t]
  \centering
   \includegraphics[width=0.85\linewidth]{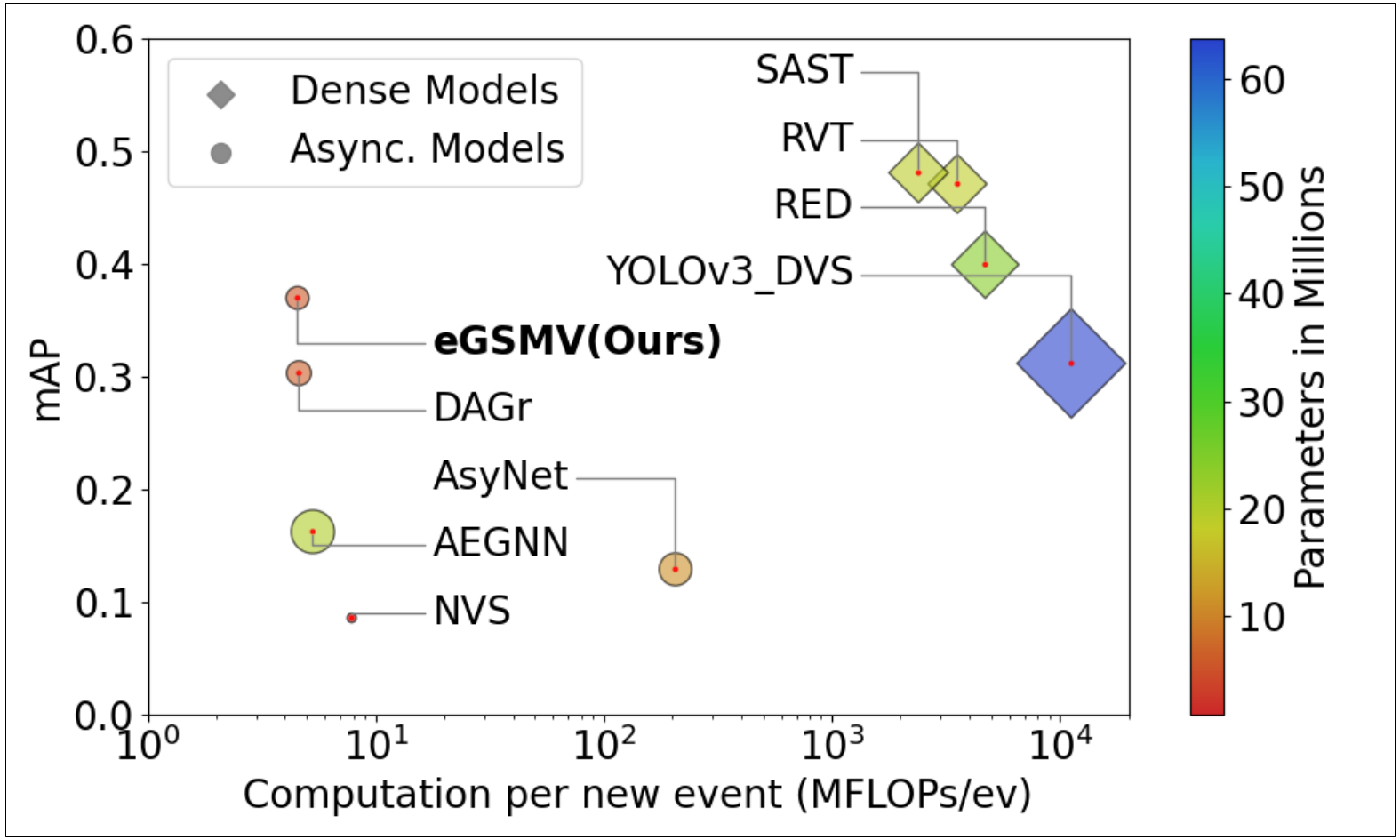}
   \caption{\textbf{Detection Summary:} Comparing performance, computation, and model size of asynchronous, dense representation-based methods on the Gen1 dataset.}
   \label{fig_methodoloy:summary}
\end{figure}

Starting with frequency-based synchronous methods, we note that Transformer+RNN models like SAST~\cite{sast}, SSM~\cite{ssm}, and RVT~\cite{rvt} consistently surpass other methods in accuracy. However, they exhibit high computational complexity, requiring an order of $1000$ times more MFLOPs per event than our method due to their dense frequency-based processing. This dense processing restricts a single event to be processed independently, contributing to their high computational cost. Moreover, as highlighted in~\cite{ssm}, RNN-based methods are sensitive to frequency changes, potentially degrading performance when the inference frequency differs from the training frequency. RED~\cite{1megapixel} shows a much narrower performance gap compared to our method, but suffers from similarly high computational demands.
We then turn to asynchronous methods, where our approach demonstrates clear advantages. Compared to state-of-the-art SNNs, particularly EAS-SNN, our model achieves comparable performance with approximately $20\%$ of its parameters. This efficiency is further emphasized when comparing \textit{eGSMV} to existing graph-based approaches. Specifically, as we can see in \figureautorefname~\ref{fig_methodoloy:summary}, our method outperforms AEGNN~\cite{aegnn} by $21\%$ while requiring only $25\%$ of its parameters. This falls in line with what we expect due to our 2D kernel design choice, which also leads to a $75\%$ lesser MFLOPs in every message passing step. Overall, the end-to-end MFLOPS/ev is only $13\%$ lesser as aggressive pooling operation is not performed in order to maintain sparsity and granularity. Against DAGr~\cite{dagr}, we evaluate two configurations. The first is without the early temporal aggregation that maintains high granularity, like our approach. We observe around a $15\%$ increase in this case. With their more optimal setting, where early aggregation to a single voxel is performed, we observe an improvement of over $6\%$. Since DAGr is an event+image fusion technique, the reported parameter size is not representative.

\begin{table}[t]
\centering
\resizebox{0.9\columnwidth}{!}{
\begin{tabular}{l|cccc}
\toprule \hline
\multicolumn{1}{c|}{\textbf{Graph Update}} & \textbf{2000} & \multicolumn{1}{c}{\textbf{4000}} & \textbf{10000} & \textbf{25000} \\ \midrule
Dense graph update  & 58.9 & 89.9 & 181.4 & 331.9 \\
Serial (SSL $\times$ MVL) & 24.4 & 25.1 & 27.4  & 29.9  \\
Parallel (SSL $\times$ MVL) & \textbf{17.7} & \textbf{18.1} & \textbf{19.9}  & \textbf{21.4} \\
\hline \bottomrule
\end{tabular}%
}
\caption{\textbf{Timing Experiments (ms) on increasing graph size:} Comparing time to process a new event by a dense graph update against our serial and parallel variants of asynchronous update.}
\label{tab:runtime}
\end{table}

We observe additional interesting insights on eTraM, which uses a static perspective. Here, the gap between dense-based methods and \textit{eGSMV} is notably smaller than on the Gen1 ego-motion dataset. This could be attributed to the high volume of background events in Gen1, which introduce noise. The close-up in the gap is observed while still requiring only a fraction of the parameters compared to other methods. We also establish benchmarks for AEGNN under the same evaluation settings as our method. In these tests, \textit{eGSMV} consistently outperforms AEGNN, achieving mAP $35\%$ higher with lower computational cost and fewer parameters. This further validates the efficiency of our approach in event-based processing tasks.

\textbf{Timing Experiments.} We compare the time our sparse and asynchronous method takes to process a new event against a dense method which updates the entire graph on an Nvidia A30 GPU (\tableautorefname~\ref{tab:runtime}). Since the SSL and MVL blocks operate on independent graphs, they can function in parallel, allowing improved efficiency.~\textit{eGSMV} needs $18.1 ms$ on a graph with $4,000$ nodes and $20.9 ms$ on a graph with $25,000$ nodes. With an increasing graph size, we observe a minimal increase in time, unlike the dense graph update method. Additionally, our quadratic kernels enable more than \textbf{5x speedup} compared to
AEGNN~\cite{aegnn}, which relies on cubic kernels. Finally, we believe that further caching optimizations like in
\cite{dagr} and implementation on suitable hardware could achieve greater improvements, as GPUs are primarily optimized for dense tensor operations.

\subsection{Ablation Studies}
This section examines each module in \textit{eGSMV} that contributes to the final result. Ablation studies are performed on the Gen1 validation set, and the results are compared against the best-performing model after $100k$ steps. To reduce the training time, each sequence has a length of 50$ms$.
More ablations are discussed in the supplementary section.

\subsubsection{Model Components}

\begin{figure*}[th]
   \centering
   \includegraphics[width=0.8\linewidth]{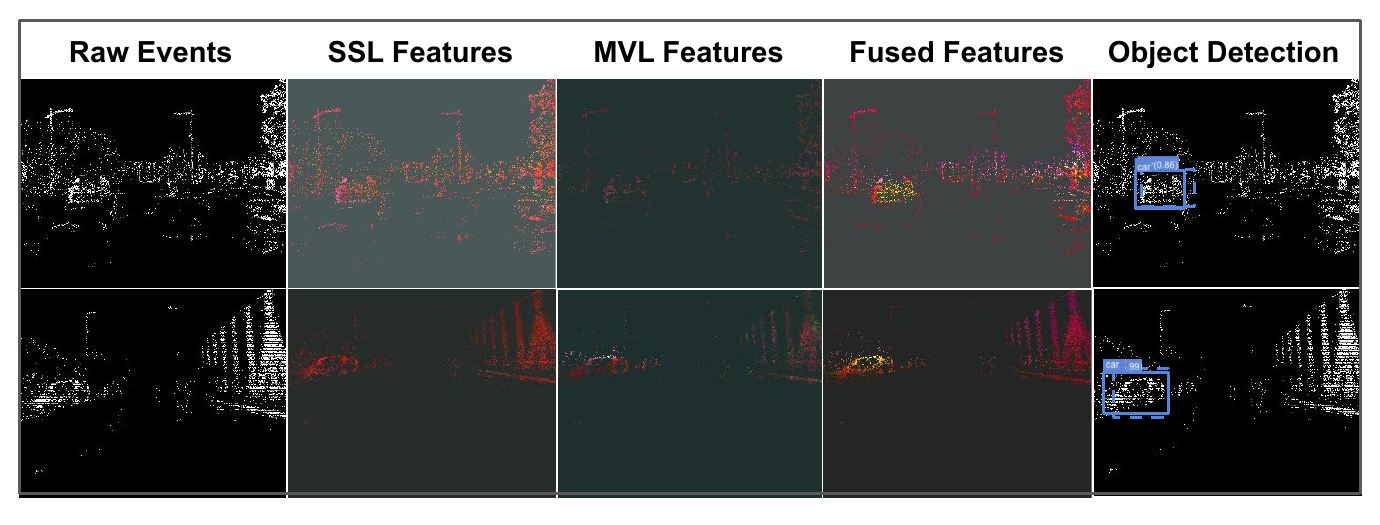}
   \caption{\textbf{Visualizations:} Qualitative illustrations of raw events, the top three principal components (from PCA) for the SSL, MVL, and the fused feature map from the final SMVL block, along with the detection results.}
    \label{fig:visualisation}
\end{figure*}

\begin{table}[t]
\centering
{\small
\begin{tabular}{ccc}
\toprule \hline
\multicolumn{1}{c}{\textbf{SSL Block}} & \textbf{MVL Block} & \textbf{mAP$\uparrow$} \\ \hline  \midrule
\multicolumn{1}{c}{\xmark}                 &           \cmark         & 0.02      \\
                        \cmark             &           \xmark         &    0.25          \\ \hline
                       \cmark                &         \cmark           &   \textbf{0.36}    
\\ \hline \bottomrule
\end{tabular}%
}
\caption{\textbf{Spatiotemporal Fusion in SMVL:} Evaluating the importance of modeling spatial as well as temporal dependencies.}
\label{tab:fusion_ablation}
\end{table}

\textbf{Spatio-Temporal Fusion.} We evaluate the impact of combining spatial and temporal features by comparing three configurations in \tableautorefname~\ref{tab:fusion_ablation}, an SSL-only model, an MVL-only model, and \textit{eGSMV} that integrates both. The MVL-only model performs significantly lower than the fused model and the SSL-only model. This is expected since a node's temporal neighbors are scarcely present in the $XY$ plane, and thus fail to understand local structures alone. While the SSL-only model performs better due to its spatial focus, it still falls short of the fused model by approximately 11\%. These findings show that even in spatial tasks like object detection, it is essential to model both spatial and temporal aspects of event data.

\textbf{Detection Head.} Here, we evaluate the impact of the event-driven granular detection head compared to the fixed grid-based heads. To examine this, we progressively coarsen the graph by max pooling event sets into voxels to achieve the target resolution. \tableautorefname~\ref{tab:detection_head} summarizes the results on varying the resolution of the feature grid. We observe a noticeable decline in performance as we increase the coarseness of the graph, showcasing the superiority of our event-driven method of detection. A small increase in performance is observed when we pool nearby neighbors to get a performance of $0.361$. This is likely due to some redundant detections leading in node-wise processing, leading to a lowered precision. 
This ablation highlights the benefits of event-driven detection for retaining granularity and enhancing detection performance, making it especially effective in node-level tasks like object detection. 

\subsection{Qualitative Comparison}
\figureautorefname~\ref{fig:visualisation} illustrates the progression of features of our model trained on Gen1 from raw event data to the final stage of object detection output. It shows how SSL and MVL features individually combine to create fused features. While the SSL features learn a global structure of the scene, the MVL features indicate the location of the dynamic object. We can further observe that the fused features are particularly active near the object with a bright yellow color. This visualization demonstrates how our method effectively leverages spatial and motion feature information while maintaining data sparsity and temporal granularity.

\begin{table}[t]
\centering
{\scriptsize
\begin{tabular}{l|llllll}
\toprule \midrule
\textbf{Voxel size} & \textbf{Node-wise} & \textbf{2x2} & \textbf{4x4} & \textbf{8x8} & \textbf{16x16} & \textbf{24x24} \\ \midrule
\textbf{mAP $\uparrow$}        & 0.353                 & \textbf{0.361}         &       0.357       & 0.341         & 0.273           & 0.197          \\
\midrule \bottomrule
\end{tabular}
}
\caption{\textbf{Granularity of Detection Head:} Analyzing the impact of pooling nodes for a fixed grid-based head.}
\label{tab:detection_head}
\end{table}

\section{Limitations and Future Work}
Despite our progress in asynchronous graph-based processing, we recognize some research areas that need further exploration beyond the scope of our paper. Importantly, unlike dense tensor-based approaches, graph-based methods face a fundamental storage bottleneck due to irregular, non-contiguous node-edge relationships, leading to inefficient memory access on hardware optimized for contiguous storage. Since most prior works, including ours, store preprocessed graphs before training, benchmarking on large-scale, high-resolution datasets like 1 Megapixel Automotive dataset~\cite{1megapixel} becomes challenging due to high memory and compute requirements. Efficient on-the-fly graph construction during training and leveraging cross-dataset knowledge transfer could be a few potential directions to tackle this. Further, we believe our representation strategy could be extrapolated to other event-level tasks as well, such as optical flow and motion understanding, where asynchronous low-latency inference could be vital. Advancing these directions would benefit from establishing widely accepted baselines and benchmarks tailored to asynchronous settings.

\section{Conclusion}
Event data, characterized by its non-Euclidean structure, sparse distribution, and asynchronous nature, poses unique challenges that traditional dense representation-based approaches struggle to address effectively.
In this regard, our work presents a novel multigraph framework that models events with separate spatial-temporal neighborhoods and enables learning quadratic relations rather than cubic. By employing an anisotropic 2D spline kernel for spatial modeling and motion vector-based attention for temporal learning, our approach outperforms the state-of-the-art graph-based method by over $6\%$, with no additional computational requirement. Through our experiments, we demonstrate that effectively modeling event graphs is key to achieving good performance. To the best of our knowledge, \textbf{\textit{eGSMV}} presents the first work at explicitly modeling event graphs to leverage its unique spatio-temporal dependencies while retaining its spatial sparsity, temporal granularity, and asynchrony. This framework demonstrates the potential of processing raw event data as a graph and, we believe, lays a strong foundation for advancing event-based vision toward more efficient, real-time applications.

\section{Acknowledgments}
This research is sponsored by NSF, the Partnerships for Innovation grant (\#2329780). We thank the Research Computing (RC) at Arizona State University (ASU) and the NSF NAIRR program for their generous support in providing computing resources. The views and opinions of the authors expressed herein do not necessarily state or reflect those of the funding agencies and employers.
\clearpage
\maketitlesupplementary

\section{Training Details}
\label{training_details}
This section outlines the training methodology for the detection task on \textit{eGSMV}. To complement the main paper, we provide additional details on the loss functions and data augmentation techniques used to optimize the graph-based data structure during training.
\subsection{Loss Modeling}
Here, we discuss the loss modeling associated with our event-driven detection head, which is inspired by YOLOX~\cite{yolox2021}. The detection head comprises a classification branch and a regression branch, each with a distinct objective and corresponding loss function.

\subsubsection{Classification branch}
The objective of the classification branch is to classify the object class of each node. Given the inherent class imbalance in the datasets and the varying scale of object classes, the number of nodes corresponding to each object class can differ significantly. To address this, a weighted cross-entropy loss is computed for each node, defined as,
\begin{equation}    
l_{cls} = - \frac{1}{N} \sum_{i=1}^{N} w_{y_{i}} \cdot y_i \cdot \log(\hat{y}_i)
\end{equation}

Here, $y_i$ is the predicted class probability, $\hat{y_i}$ is the one-hot vector of the ground truth class and $w_{y_i}$ is a constant weight assigned to the ground truth class of node $v_i$ to account for the class imbalance in the dataset.

\subsubsection{Regression Branch}
The regression branch is responsible for predicting the relative bounding box coordinates and the Intersection over Union (IoU) confidence, $s$. Losses are only considered if the node is a non-background event. To improve localization accuracy and penalize incorrect bounding box dimensions, we use cIoU loss~\cite{ciou} and Huber loss~\cite{huberloss}, computed as follows,

\begin{equation}
l_{\text{loc}} = \frac{1}{\sum_{i \in \mathbb{N}} \mathbf{1}_{\{c_i \neq \text{bg}\}}} \sum_{i \in \mathbb{N}} \mathbf{1}_{\{c_i \neq \text{bg}\}} \cdot \ell_{ciou}(x_i, \hat{x_i})
\end{equation}

\begin{equation}
l_{\text{dim}} = \frac{1}{\sum_{i \in \mathbb{N}} \mathbf{1}_{\{c_i \neq \text{bg}\}}} \sum_{i \in \mathbb{N}} \mathbf{1}_{\{c_i \neq \text{bg}\}} \cdot \ell_{huber}(x^{wh}_i, \hat{x^{wh}_i})
\end{equation}

Here, $x_i$ represents the bounding box prediction for node $v_i$, $\hat{x_i}$ is the corresponding ground truth, and $x^{wh}$ refers to the predicted width and height. The losses are computed only for the nodes with ground truth class $c_i \neq bg$, where $bg$ refers to the background class.

Additionally, we compute the binary cross-entropy on the IoU confidence score to evaluate the confidence of each bounding box prediction. The confidence score is designed to be low if the IoU between the predicted bounding box and the ground truth is less than $0.5$. The confidence loss is defined as,
\begin{equation}
l_{conf} = - \frac{1}{N} \sum_{i=1}^{N} \left[ s_i \log(\hat{s}_i) + (1 - s_i) \log(1 - \hat{s}_i) \right]
\end{equation}

where predicted confidence score $\hat{s_i}(x_i, \hat{x_i})$ is computed as,
\begin{equation}
\hat{s_i}(x_i, \hat{x_i}) = 
\begin{cases} 
1 & \text{if IoU}(x_i, \hat{x_i}) \geq 0.5 \\
0 & \text{otherwise}
\end{cases}
\end{equation}

Here, $s_i$ represents the IoU confidence score of the bounding box prediction for node $v_i$. 

\subsubsection{Total loss}
The total loss is a weighted summation of all individual losses,
\begin{equation}
l_{total} = \alpha l_{cls} + \beta l_{loc} + \gamma l_{dim} + \lambda l_{conf}
\end{equation}

The loss weights are set as $\alpha=1$, $\beta=2$, $\gamma=3$, and $\lambda=1.5$.

\begin{table}[t]
\centering
{\small
\begin{tabular}{lccc} \toprule \midrule
             &             & \multicolumn{2}{l}{\textbf{Magnitude}} \\ \cline{3-4}
\textbf{Augmentation} & \textbf{Probability} & \textbf{min}           & \textbf{max}           \\ \midrule
Translation  & 0.5         & 0.05          & 0.15          \\
Cropping     & 0.4         & 0.05          & 0.25           \\ \midrule \bottomrule         
\end{tabular}
}
\caption{\textbf{Data Augmentation} Probability and range of the magnitude for the application of translation and cropping augmentations.}
\label{tab:data_aug}
\end{table}

\subsection{Data Augmentation}
We apply two types of data augmentations to train our model from scratch, with parameters summarized in \tableautorefname~\ref{tab:data_aug}. Since each training sequence must contain at least one bounding box, we randomly choose a bounding box as the anchor box and perform the augmentations around it.

The first augmentation performed is a random translation along the $x$ and $y$ coordinates, applied with a probability of $0.5$. The maximum translation in each dimension is restricted between $5\%$ and $15\%$ of the input shape. The second augmentation is random cropping around the anchor bounding box with the probability of cropping set at $0.35$. The cropping size is constrained to a minimum of $5\%$ and a maximum of $25\%$ of the input dimensions. 

\section{Network Architecture}
The initial node features $\mathbf{x}$ are projected into a $16$-dimensional space using an MLP$(4,16)$. Each subsequent layer of the network processes the node features through an SMVL block, which combines features from the SSL and MVL components. At layer $\mathbf{n}$, the input features of shape $M^\mathbf{n}_{in}$ are transformed into a spatially and temporally aware node feature of shape $M^\mathbf{n}_{out}$. The network outputs have channels $M_{out}=(16, 16, 32, 32, 64, 64, 128, 128)$.  The SSL block has a kernel size of $(8,8,1)$ and the MVL block has $4$ heads in the backbone. For the detection head, the SSL block has a kernel size of $(5,5,1)$ with $1$ head in the MVL block, which downsamples node features to a $64$-dimensional shape.
Batch normalization and ReLU activations are applied after each step but omitted in illustrations for conciseness.

\begin{table}[t]
\centering
{\small
\begin{tabular}{lcc}
\toprule \hline
\multicolumn{1}{c}{\textbf{SSL Block}} & \multicolumn{1}{c}{\textbf{mAP $\uparrow$}} & \textbf{Params (M)$\downarrow$} \\ \hline \midrule
GCN                      &   0.09  & \textbf{4.7}\\
3D Isotropic SplineConv   & \textbf{0.38} & 34.9 \\ \hline
2D Anisotropic SplineConv & 0.36 & 5.6\\
\hline \bottomrule
\end{tabular}%
}
\caption{\textbf{Spatial Structure Learning:} 3D kernels provide a marginal performance improvement at a huge parameter expense.}
\label{tab:ssl_block}
\end{table}

\section{Additional Experiments}
While the main paper focused on experiments validating the importance of fusion and the impact of the detection head granularity in \textit{eGSMV}, this section investigates each component of the SMVL block and the impact of other parameters from graph construction on the performance of our framework.
\subsection{Ablation on Model Components}
\textbf{SSL Block.} Here, we evaluate the impact of various spatial structure learning techniques for the SSL block. Presented in \tableautorefname~\ref{tab:ssl_block}, the configurations compared are a standard GCN layer, an isotropic 3D spline kernel, and our proposed anisotropic 2D spline kernel. The MVL block is kept constant across all models for a fair comparison.

GCN serves as a baseline, aggregating spatial neighbors uniformly without spatial adaptiveness. The isotropic 3D kernel, similar to the approach used in previous works, introduces additional temporal depth, leading to an increased parameter count and computational overhead. Results indicate that both spline-based kernels outperform the GCN-based SSL. This could be attributed to GCN being more prone to overfitting than its Spline variants. Further, the 3D isotropic kernel adds a substantial number of parameters with only a marginal performance gain over the 2D variant. This suggests that while the spline kernel does well at capturing spatial features, the additional temporal context in the 3D setup provides limited benefit.

\begin{table}[t]
\centering
{\small
\begin{tabular}{lcc}
\toprule \hline
\multicolumn{1}{c}{\textbf{MVL Block}} & \multicolumn{1}{c}{\textbf{mAP $\uparrow$}} & \textbf{Params (M) $\downarrow$} \\ \hline \midrule
w/o motion vector features  & 0.33 & 5.6 \\
w motion vector features (ours)   & \textbf{0.36} & 5.6 \\
\hline \bottomrule
\end{tabular}%
}
\caption{\textbf{Motion Vector Learning:} Motion guidance improves performance in attention-based temporal learning.}
\label{tab:mvl_block}
\end{table}

\textbf{MVL Block.} This ablation examines the impact of incorporating motion vector features within the MVL block. In Table~\ref{tab:mvl_block}, we compare two models: one with motion vector features encoded in the edge and one without. These results indicate that adding motion vector features improves mAP by 3\% with a negligible increase in model size, demonstrating that motion vector guidance effectively enhances temporal representation without significant computational cost.

Next, we also evaluate the impact of multi-head attention with uniform aggregation as well as its single-head counterpart. As demonstrated through Table~\ref{tab:mvl_mha}, multi-head attention (MHA) in MVL enables a different weighting to temporal neighbors that carry varying temporal motion cues due to differences in time and their spatial location. Temporal neighbors carry cues with lower inductive bias, unlike spatial neighbors. 

\begin{table}[t]
\centering
{\small
\begin{tabular}{ll}
\toprule \hline
\multicolumn{1}{c}{\textbf{Temporal aggregation}} & \multicolumn{1}{c}{\textbf{mAP $\uparrow$}}\\ \hline \midrule
\multicolumn{1}{l}{Uniform aggregation} & \multicolumn{1}{l}{0.31} \\
Single-head attention               & 0.34                    \\ \hline
Multi-head attention (ours)               & \textbf{0.36}                    \\
\hline \bottomrule
\end{tabular}
}
\caption{\textbf{Multihead attention:} Evaluating impact of different aggregation methods in the MVL block.}
\label{tab:mvl_block}
\end{table}

\begin{figure*}[t]
  \centering
   \includegraphics[width=0.88\linewidth]{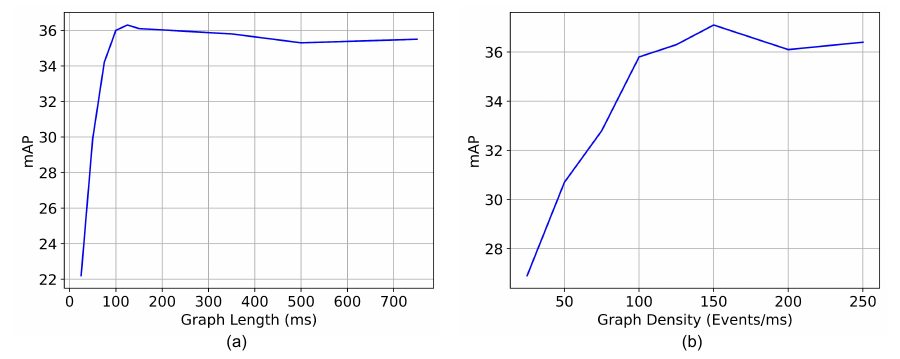}
    \caption{\textbf{Impact of Graph Construction on Performance.} (a) Variation in mAP with increasing graph length sequences, showing performance improvement up to an optimal sequence length before plateauing; (b) Variation in mAP with increasing graph density (number of events per ms), showing how performance improves with density up to a point before saturating.}
   \label{fig:supp_exp}
\end{figure*}

\subsection{Impact of Graph Length}
In this experiment, we analyze the impact of graph length, defined as the time window, on the model's ability to capture spatial and temporal dependencies and, consequently, detection performance.  A longer graph length accumulates more temporal context at the cost of increased computational requirements. However, a shorter graph with a very small look-back can miss critical temporal dynamics, particularly crucial in detecting slow-moving objects.

To evaluate this, we test the performance of our method on graph length sequences ranging from $25ms$ to $750ms$ as illustrated in \figureautorefname~\ref{fig:supp_exp}(a). We observe that graphs of a smaller sequence length underperform due to insufficient temporal context. The model achieves peak performance at a graph length of $100ms$, beyond which we observe diminishing returns in detection accuracy. This suggests that a look-back of $100ms$ strikes the optimal balance, capturing sufficient temporal information.

\begin{figure*}[t]
  \centering
   \includegraphics[width=0.9\linewidth]{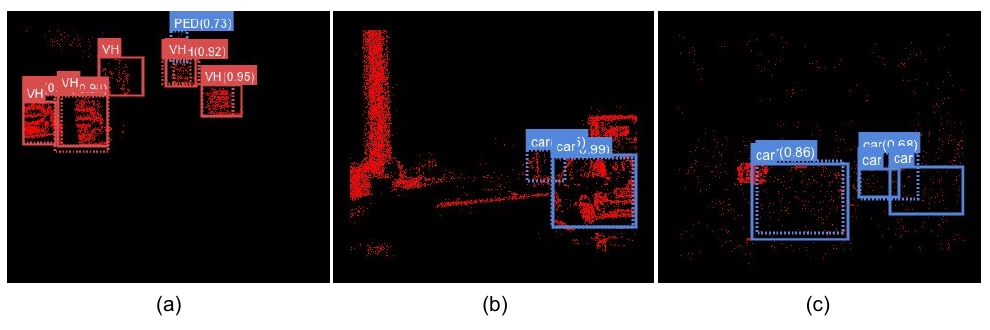}
   \caption{\textbf{Object Detection by eGSMV.} (a) Detection results on the eTraM dataset showcasing high event density, (b) Detection from the Gen1 dataset with dynamic motion, and (c) Detection from the Gen1 dataset in a stationary scenario with sparse events.}
   \label{fig:qualitative_detections}
\end{figure*}

\begin{figure*}[t]
  \centering
   \includegraphics[width=0.95\linewidth]{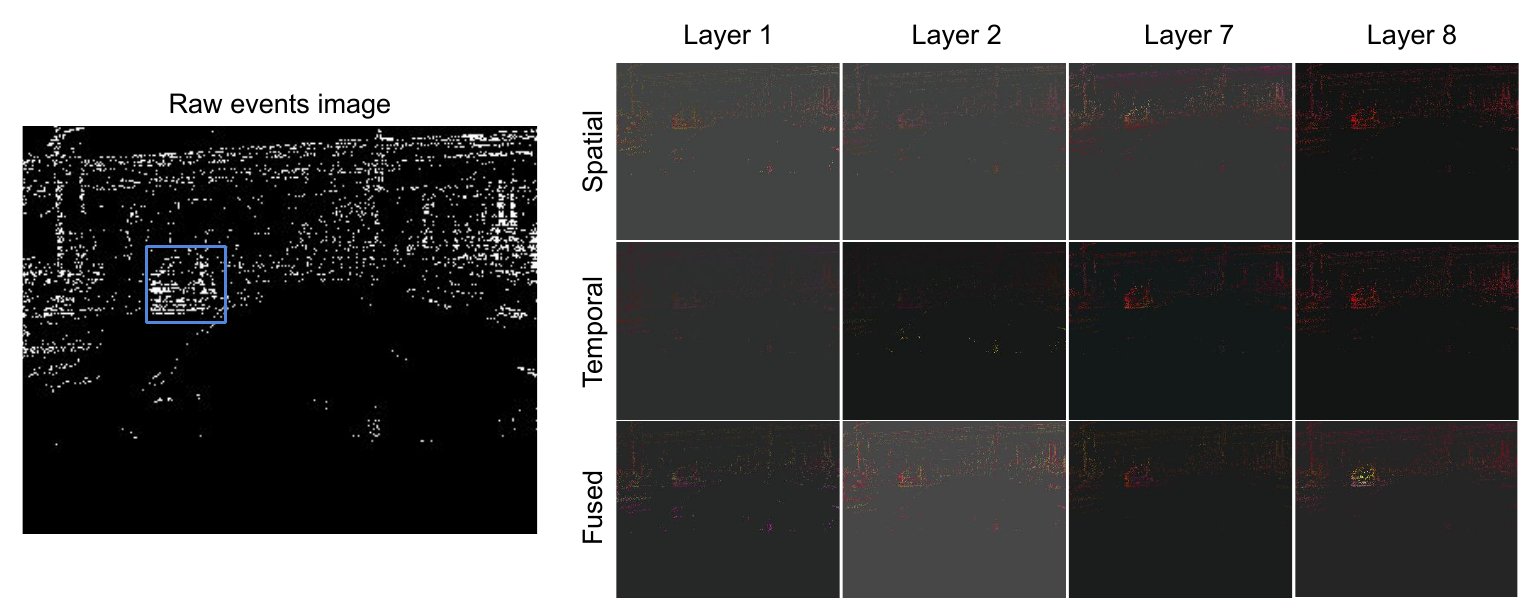}
    \includegraphics[width=0.95\linewidth]{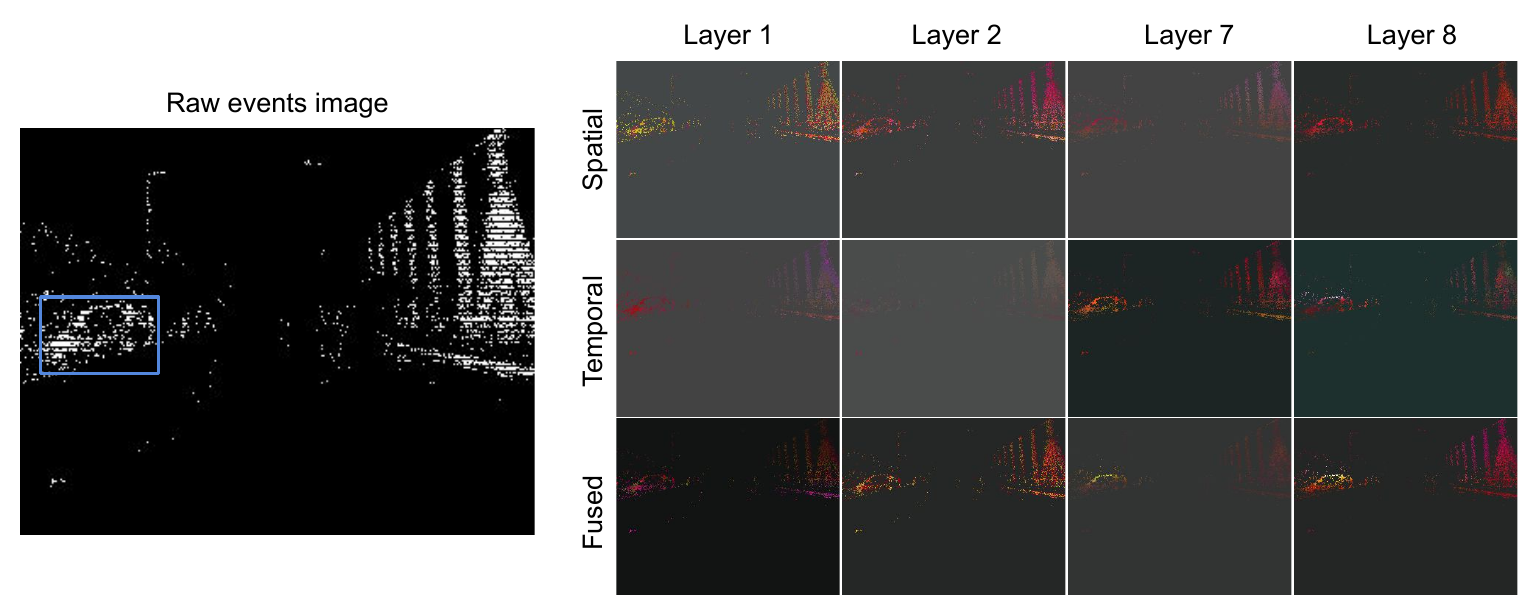}
   \caption{\textbf{Feature Maps from Inference.} Visualizations of raw events alongside the top three principal components (PCA) of the spatial (SSL), temporal (MVL), and fused feature maps from the first two and final two SMVL layers.}
   \label{fig:qualitative_maps}
\end{figure*}

\subsection{Impact of Graph Density}
Graph density, determined by the number of nodes in a $1ms$ time window, directly influences the model's ability to aggregate meaningful features and manage computational requirements. This motivates our study of sampling density and its effect on performance. Aggressive sampling, which limits the number of nodes, may result in insufficient spatial and temporal interactions and reduce the model's ability to learn robust features. On the other hand, a denser graph provides more context per node, potentially enhancing spatial and temporal feature learning. However, excessive density increases computational complexity and memory requirements, potentially leading to overfitting.

We present a systematic variation in graph density by adjusting the number of permissible events per $1ms$ window from $25$ to $250$ as illustrated in \figureautorefname~\ref{fig:supp_exp}(b). We observe that moderately dense graphs achieve the best trade-off, enabling robust feature learning while having the least computational burden. This finding shows the importance of carefully tuning graph density to optimize performance in event-based vision tasks.

\section{Additional Visualizations}
We present qualitative results highlighting the detection performance across different datasets and scenarios, as shown in \figureautorefname~\ref{fig:qualitative_detections}. The visualizations provide insights into how \textit{eGSMV} performs in different event distributions, motion dynamics, and background activity levels.

In \figureautorefname~\ref{fig:qualitative_detections}(a), detection results from the eTraM dataset demonstrate the localization capabilities in sequences with a sparse event distribution sequences and minimal background events. \figureautorefname~\ref{fig:qualitative_detections}(b) showcases the ability to capture spatiotemporal dependencies for accurate detection in high background activity cases when there is dynamic motion involved in the Gen1 dataset. Finally,
\figureautorefname~\ref{fig:qualitative_detections}(c) presents detection results from the Gen1 dataset in a sparse event distribution under stationary conditions. Despite the challenges posed by limited event generation, the model, albeit with reduced confidence levels, successfully localizes the objects. These visualizations contain events from the most recent $25ms$.

\figureautorefname~\ref{fig:qualitative_maps} provides a detailed visualization of how the architecture progressively localizes relevant objects in the scene across layers. The figure presents PCA-based feature maps for SSL, MVL, and fused representations from the initial and final SMVL layers. Initially, in the earlier layers, a majority of the events are assigned high importance, capturing broad spatial and temporal features. As the network progresses to deeper layers, the architecture refines its focus, selectively emphasizing events corresponding to the objects in the scene. As highlighted by these visualizations, this hierarchical learning process demonstrates the model's ability to integrate spatial and temporal dependencies effectively, allowing it to localize objects with increasing precision at deeper layers.

\section{Dataset Licenses}
\textbf{Gen1}~\cite{gen1}
``Prophesee Gen1 Automotive Detection Dataset License Terms and Conditions": \url{https://www.prophesee.ai/2020/01/24/prophesee-gen1- automotive-detection-dataset/}
\\
\\
\textbf{eTraM}~\cite{verma2024etram}
``Creative Commons Attribution-ShareAlike 4.0 International License." \url{https://github.com/eventbasedvision/eTraM}

{
    \small
    \bibliographystyle{ieeenat_fullname}
    \bibliography{main}

\begin{thebibliography}{61}
\providecommand{\natexlab}[1]{#1}
\providecommand{\url}[1]{\texttt{#1}}
\expandafter\ifx\csname urlstyle\endcsname\relax
  \providecommand{\doi}[1]{doi: #1}\else
  \providecommand{\doi}{doi: \begingroup \urlstyle{rm}\Url}\fi

\bibitem[Aliminati et~al.(2024)Aliminati, Chakravarthi, Verma, Vaghela, Wei, Zhou, and Yang]{aliminati2024sevd}
Manideep~Reddy Aliminati, Bharatesh Chakravarthi, Aayush~Atul Verma, Arpitsinh Vaghela, Hua Wei, Xuesong Zhou, and Yezhou Yang.
\newblock Sevd: Synthetic event-based vision dataset for ego and fixed traffic perception.
\newblock \emph{arXiv preprint arXiv:2404.10540}, 2024.

\bibitem[Amir et~al.(2017)Amir, Taba, Berg, Melano, McKinstry, Di~Nolfo, Nayak, Andreopoulos, Garreau, Mendoza, Kusnitz, Debole, Esser, Delbruck, Flickner, and Modha]{low_power}
Arnon Amir, Brian Taba, David Berg, Timothy Melano, Jeffrey McKinstry, Carmelo Di~Nolfo, Tapan Nayak, Alexander Andreopoulos, Guillaume Garreau, Marcela Mendoza, Jeff Kusnitz, Michael Debole, Steve Esser, Tobi Delbruck, Myron Flickner, and Dharmendra Modha.
\newblock A low power, fully event-based gesture recognition system.
\newblock In \emph{2017 IEEE Conference on Computer Vision and Pattern Recognition (CVPR)}, pages 7388--7397, 2017.

\bibitem[Bhattacharya et~al.()Bhattacharya, Cannici, Rao, Tao, Kumar, Matni, and Scaramuzza]{robotics_event}
Anish Bhattacharya, Marco Cannici, Nishanth Rao, Yuezhan Tao, Vijay Kumar, Nikolai Matni, and Davide Scaramuzza.
\newblock Monocular event-based vision for obstacle avoidance with a quadrotor.
\newblock In \emph{8th Annual Conference on Robot Learning}.

\bibitem[Bi et~al.(2019)Bi, Chadha, Abbas, Bourtsoulatze, and Andreopoulos]{bi2019graph}
Yin Bi, Aaron Chadha, Alhabib Abbas, Eirina Bourtsoulatze, and Yiannis Andreopoulos.
\newblock Graph-based object classification for neuromorphic vision sensing.
\newblock In \emph{Proceedings of the IEEE/CVF international conference on computer vision}, pages 491--501, 2019.

\bibitem[Bi et~al.(2020)Bi, Chadha, Abbas, Bourtsoulatze, and Andreopoulos]{graph1}
Yin Bi, Aaron Chadha, Alhabib Abbas, Eirina Bourtsoulatze, and Yiannis Andreopoulos.
\newblock Graph-based spatio-temporal feature learning for neuromorphic vision sensing.
\newblock \emph{IEEE Transactions on Image Processing}, 2020.

\bibitem[Bolten et~al.(2021)Bolten, Pohle-Fröhlich, and Tönnies]{dvsoutlab}
Tobias Bolten, Regina Pohle-Fröhlich, and Klaus~D. Tönnies.
\newblock Dvs-outlab: A neuromorphic event-based long time monitoring dataset for real-world outdoor scenarios.
\newblock In \emph{2021 IEEE/CVF Conference on Computer Vision and Pattern Recognition Workshops (CVPRW)}, pages 1348--1357, 2021.

\bibitem[Boretti et~al.(2023)Boretti, Bich, Pareschi, Prono, Rovatti, and Setti]{pedro}
Chiara Boretti, Philippe Bich, Fabio Pareschi, Luciano Prono, Riccardo Rovatti, and Gianluca Setti.
\newblock Pedro: An event-based dataset for person detection in robotics.
\newblock In \emph{Proceedings of the IEEE/CVF Conference on Computer Vision and Pattern Recognition}, pages 4064--4069, 2023.

\bibitem[Brody et~al.()Brody, Alon, and Yahav]{gatv2}
Shaked Brody, Uri Alon, and Eran Yahav.
\newblock How attentive are graph attention networks?
\newblock In \emph{International Conference on Learning Representations}.

\bibitem[Cannici et~al.(2019)Cannici, Ciccone, Romanoni, and Matteucci]{conv2019async}
Marco Cannici, Marco Ciccone, Andrea Romanoni, and Matteo Matteucci.
\newblock Asynchronous convolutional networks for object detection in neuromorphic cameras.
\newblock In \emph{Proceedings of the IEEE/CVF Conference on Computer Vision and Pattern Recognition Workshops}, pages 0--0, 2019.

\bibitem[Chakravarthi et~al.(2024)Chakravarthi, Verma, Daniilidis, Fermuller, and Yang]{chakravarthi2024recent}
Bharatesh Chakravarthi, Aayush~Atul Verma, Kostas Daniilidis, Cornelia Fermuller, and Yezhou Yang.
\newblock Recent event camera innovations: A survey.
\newblock \emph{arXiv preprint arXiv:2408.13627}, 2024.

\bibitem[Cordone et~al.(2022{\natexlab{a}})Cordone, Miramond, and Thierion]{snn3}
Lo{\"\i}c Cordone, Beno{\^\i}t Miramond, and Philippe Thierion.
\newblock Object detection with spiking neural networks on automotive event data.
\newblock In \emph{2022 International Joint Conference on Neural Networks (IJCNN)}, 2022{\natexlab{a}}.

\bibitem[Cordone et~al.(2022{\natexlab{b}})Cordone, Miramond, and Thierion]{vc-densenet}
Lo{\"\i}c Cordone, Beno{\^\i}t Miramond, and Philippe Thierion.
\newblock Object detection with spiking neural networks on automotive event data.
\newblock In \emph{2022 International Joint Conference on Neural Networks (IJCNN)}, pages 1--8. IEEE, 2022{\natexlab{b}}.

\bibitem[de~Tournemire et~al.(2020)de~Tournemire, Nitti, Perot, Migliore, and Sironi]{gen1}
Pierre de Tournemire, Davide Nitti, Etienne Perot, Davide Migliore, and Amos Sironi.
\newblock A large scale event-based detection dataset for automotive, 2020.

\bibitem[Deng et~al.(2022)Deng, Chen, Liu, and Li]{graph4}
Yongjian Deng, Hao Chen, Hai Liu, and Youfu Li.
\newblock A voxel graph cnn for object classification with event cameras.
\newblock In \emph{Proceedings of the IEEE/CVF Conference on Computer Vision and Pattern Recognition}, pages 1172--1181, 2022.

\bibitem[Deng et~al.(2024)Deng, Chen, and Li]{edgcn}
Yongjian Deng, Hao Chen, and Youfu Li.
\newblock A dynamic gcn with cross-representation distillation for event-based learning.
\newblock \emph{Proceedings of the AAAI Conference on Artificial Intelligence}, 38\penalty0 (2):\penalty0 1492--1500, 2024.

\bibitem[Falcon and {The PyTorch Lightning team}(2019)]{lightning}
William Falcon and {The PyTorch Lightning team}.
\newblock {PyTorch Lightning}, 2019.

\bibitem[Fey and Lenssen(2019)]{pyg}
Matthias Fey and Jan~E. Lenssen.
\newblock Fast graph representation learning with {PyTorch Geometric}.
\newblock In \emph{ICLR Workshop on Representation Learning on Graphs and Manifolds}, 2019.

\bibitem[Gallego et~al.(2017)Gallego, Lund, Mueggler, Rebecq, Delbruck, and Scaramuzza]{filter3}
Guillermo Gallego, Jon~EA Lund, Elias Mueggler, Henri Rebecq, Tobi Delbruck, and Davide Scaramuzza.
\newblock Event-based, 6-dof camera tracking from photometric depth maps.
\newblock \emph{IEEE transactions on pattern analysis and machine intelligence}, 40\penalty0 (10):\penalty0 2402--2412, 2017.

\bibitem[Gallego et~al.(2019)Gallego, Delbr{\"{u}}ck, Orchard, Bartolozzi, Taba, Censi, Leutenegger, Davison, Conradt, Daniilidis, and Scaramuzza]{eventvisionsurvey}
Guillermo Gallego, Tobi Delbr{\"{u}}ck, Garrick Orchard, Chiara Bartolozzi, Brian Taba, Andrea Censi, Stefan Leutenegger, Andrew~J. Davison, J{\"{o}}rg Conradt, Kostas Daniilidis, and Davide Scaramuzza.
\newblock Event-based vision: {A} survey.
\newblock \emph{CoRR}, abs/1904.08405, 2019.

\bibitem[Ge et~al.(2021)Ge, Liu, Wang, Li, and Sun]{yolox2021}
Zheng Ge, Songtao Liu, Feng Wang, Zeming Li, and Jian Sun.
\newblock Yolox: Exceeding yolo series in 2021.
\newblock \emph{arXiv preprint arXiv:2107.08430}, 2021.

\bibitem[Gehrig and Scaramuzza(2024)]{dagr}
Daniel Gehrig and Davide Scaramuzza.
\newblock Low-latency automotive vision with event cameras.
\newblock \emph{Nature}, 629\penalty0 (8014):\penalty0 1034--1040, 2024.

\bibitem[Gehrig and Scaramuzza(2023)]{rvt}
Mathias Gehrig and Davide Scaramuzza.
\newblock Recurrent vision transformers for object detection with event cameras.
\newblock In \emph{Proceedings of the IEEE/CVF conference on computer vision and pattern recognition}, pages 13884--13893, 2023.

\bibitem[Gehrig et~al.(2021)Gehrig, Aarents, Gehrig, and Scaramuzza]{dsec}
Mathias Gehrig, Willem Aarents, Daniel Gehrig, and Davide Scaramuzza.
\newblock Dsec: A stereo event camera dataset for driving scenarios.
\newblock \emph{IEEE Robotics and Automation Letters}, 2021.

\bibitem[Girshick et~al.(2015)Girshick, Donahue, Darrell, and Malik]{rcnn}
Ross Girshick, Jeff Donahue, Trevor Darrell, and Jitendra Malik.
\newblock Region-based convolutional networks for accurate object detection and segmentation.
\newblock \emph{IEEE transactions on pattern analysis and machine intelligence}, 38\penalty0 (1):\penalty0 142--158, 2015.

\bibitem[Hamaguchi et~al.(2023)Hamaguchi, Furukawa, Onishi, and Sakurada]{hmnet}
Ryuhei Hamaguchi, Yasutaka Furukawa, Masaki Onishi, and Ken Sakurada.
\newblock Hierarchical neural memory network for low latency event processing.
\newblock In \emph{Proceedings of the IEEE/CVF Conference on Computer Vision and Pattern Recognition}, pages 22867--22876, 2023.

\bibitem[Huber(1992)]{huberloss}
Peter~J Huber.
\newblock Robust estimation of a location parameter.
\newblock In \emph{Breakthroughs in statistics: Methodology and distribution}, pages 492--518. Springer, 1992.

\bibitem[Iacono et~al.(2018)Iacono, Weber, Glover, and Bartolozzi]{inception_ssd}
Massimiliano Iacono, Stefan Weber, Arren Glover, and Chiara Bartolozzi.
\newblock Towards event-driven object detection with off-the-shelf deep learning.
\newblock In \emph{2018 IEEE/RSJ International Conference on Intelligent Robots and Systems (IROS)}, pages 1--9, 2018.

\bibitem[Jeziorek et~al.(2023)Jeziorek, Pinna, and Kryjak]{graph5}
Kamil Jeziorek, Andrea Pinna, and Tomasz Kryjak.
\newblock Memory-efficient graph convolutional networks for object classification and detection with event cameras.
\newblock In \emph{2023 Signal Processing: Algorithms, Architectures, Arrangements, and Applications (SPA)}, pages 160--165, 2023.

\bibitem[Jiang et~al.(2019{\natexlab{a}})Jiang, Xia, Huang, Stechele, Chen, Bing, and Knoll]{Jiang2019MixedFF}
Zhuangyi Jiang, Pengfei Xia, Kai Huang, Walter Stechele, G. Chen, Zhenshan Bing, and Alois Knoll.
\newblock Mixed frame-/event-driven fast pedestrian detection.
\newblock \emph{2019 International Conference on Robotics and Automation (ICRA)}, pages 8332--8338, 2019{\natexlab{a}}.

\bibitem[Jiang et~al.(2019{\natexlab{b}})Jiang, Xia, Huang, Stechele, Chen, Bing, and Knoll]{yolov3_events}
Zhuangyi Jiang, Pengfei Xia, Kai Huang, Walter Stechele, Guang Chen, Zhenshan Bing, and Alois Knoll.
\newblock Mixed frame-/event-driven fast pedestrian detection.
\newblock In \emph{2019 International Conference on Robotics and Automation (ICRA)}, pages 8332--8338, 2019{\natexlab{b}}.

\bibitem[Kim et~al.(2016)Kim, Leutenegger, and Davison]{filter2}
Hanme Kim, Stefan Leutenegger, and Andrew~J. Davison.
\newblock Real-time 3d reconstruction and 6-dof tracking with an event camera.
\newblock In \emph{European Conference on Computer Vision}, 2016.

\bibitem[Lagorce et~al.(2017)Lagorce, Orchard, Galluppi, Shi, and Benosman]{HOTS}
Xavier Lagorce, Garrick Orchard, Francesco Galluppi, Bertram~E. Shi, and Ryad~B. Benosman.
\newblock Hots: A hierarchy of event-based time-surfaces for pattern recognition.
\newblock \emph{IEEE Transactions on Pattern Analysis and Machine Intelligence}, 39\penalty0 (7):\penalty0 1346--1359, 2017.

\bibitem[Li et~al.(2022)Li, Li, Zhu, Xiang, Huang, and Tian]{astmnet}
Jianing Li, Jia Li, Lin Zhu, Xijie Xiang, Tiejun Huang, and Yonghong Tian.
\newblock Asynchronous spatio-temporal memory network for continuous event-based object detection.
\newblock \emph{IEEE Transactions on Image Processing}, 2022.

\bibitem[Li et~al.(2021{\natexlab{a}})Li, Zhou, Yang, Zhang, Cui, Bao, and Zhang]{graph3}
Yijin Li, Han Zhou, Bangbang Yang, Ye Zhang, Zhaopeng Cui, Hujun Bao, and Guofeng Zhang.
\newblock Graph-based asynchronous event processing for rapid object recognition.
\newblock In \emph{Proceedings of the IEEE/CVF International Conference on Computer Vision}, pages 934--943, 2021{\natexlab{a}}.

\bibitem[Li et~al.(2021{\natexlab{b}})Li, Zhou, Yang, Zhang, Cui, Bao, and Zhang]{nvs}
Yijin Li, Han Zhou, Bangbang Yang, Ye Zhang, Zhaopeng Cui, Hujun Bao, and Guofeng Zhang.
\newblock Graph-based asynchronous event processing for rapid object recognition.
\newblock In \emph{Proceedings of the IEEE/CVF International Conference on Computer Vision}, pages 934--943, 2021{\natexlab{b}}.

\bibitem[Lichtsteiner et~al.(2008)Lichtsteiner, Posch, and Delbruck]{latency_async}
Patrick Lichtsteiner, Christoph Posch, and Tobi Delbruck.
\newblock A 128$\times$ 128 120 db 15 $\mu$s latency asynchronous temporal contrast vision sensor.
\newblock \emph{IEEE Journal of Solid-State Circuits}, 43\penalty0 (2):\penalty0 566--576, 2008.

\bibitem[Lin et~al.(2014)Lin, Maire, Belongie, Hays, Perona, Ramanan, Doll{\'a}r, and Zitnick]{coco}
Tsung-Yi Lin, Michael Maire, Serge Belongie, James Hays, Pietro Perona, Deva Ramanan, Piotr Doll{\'a}r, and C~Lawrence Zitnick.
\newblock Microsoft coco: Common objects in context.
\newblock In \emph{Computer Vision--ECCV 2014: 13th European Conference, Zurich, Switzerland, September 6-12, 2014, Proceedings, Part V 13}, pages 740--755. Springer, 2014.

\bibitem[Messikommer et~al.(2020{\natexlab{a}})Messikommer, Gehrig, Loquercio, and Scaramuzza]{asynet}
Nico Messikommer, Daniel Gehrig, Antonio Loquercio, and Davide Scaramuzza.
\newblock Event-based asynchronous sparse convolutional networks.
\newblock In \emph{Computer Vision--ECCV 2020: 16th European Conference, Glasgow, UK, August 23--28, 2020, Proceedings, Part VIII 16}, pages 415--431. Springer, 2020{\natexlab{a}}.

\bibitem[Messikommer et~al.(2020{\natexlab{b}})Messikommer, Gehrig, Loquercio, and Scaramuzza]{eventscn}
Nico Messikommer, Daniel Gehrig, Antonio Loquercio, and Davide Scaramuzza.
\newblock Event-based asynchronous sparse convolutional networks.
\newblock In \emph{Computer Vision--ECCV 2020: 16th European Conference, Glasgow, UK, August 23--28, 2020, Proceedings, Part VIII 16}, pages 415--431. Springer, 2020{\natexlab{b}}.

\bibitem[Mitrokhin et~al.(2020)Mitrokhin, Hua, Fermüller, and Aloimonos]{graph_segment}
Anton Mitrokhin, Zhiyuan Hua, Cornelia Fermüller, and Yiannis Aloimonos.
\newblock Learning visual motion segmentation using event surfaces.
\newblock In \emph{2020 IEEE/CVF Conference on Computer Vision and Pattern Recognition (CVPR)}, pages 14402--14411, 2020.

\bibitem[Peng et~al.(2023{\natexlab{a}})Peng, Zhang, Xiao, Sun, and Wu]{peng2023better}
Yansong Peng, Yueyi Zhang, Peilin Xiao, Xiaoyan Sun, and Feng Wu.
\newblock Better and faster: Adaptive event conversion for event-based object detection.
\newblock In \emph{Proceedings of the AAAI Conference on Artificial Intelligence}, pages 2056--2064, 2023{\natexlab{a}}.

\bibitem[Peng et~al.(2023{\natexlab{b}})Peng, Zhang, Xiong, Sun, and Wu]{get-t}
Yansong Peng, Yueyi Zhang, Zhiwei Xiong, Xiaoyan Sun, and Feng Wu.
\newblock Get: Group event transformer for event-based vision.
\newblock In \emph{International Conference on Computer Vision (ICCV)}, 2023{\natexlab{b}}.

\bibitem[Peng et~al.(2024)Peng, Li, Zhang, Sun, and Wu]{sast}
Yansong Peng, Hebei Li, Yueyi Zhang, Xiaoyan Sun, and Feng Wu.
\newblock Scene adaptive sparse transformer for event-based object detection.
\newblock In \emph{Proceedings of the IEEE/CVF Conference on Computer Vision and Pattern Recognition (CVPR)}, pages 16794--16804, 2024.

\bibitem[Perot et~al.(2020)Perot, de~Tournemire, Nitti, Masci, and Sironi]{1megapixel}
Etienne Perot, Pierre de Tournemire, Davide Nitti, Jonathan Masci, and Amos Sironi.
\newblock Learning to detect objects with a 1 megapixel event camera.
\newblock In \emph{Advances in Neural Information Processing Systems}, pages 16639--16652. Curran Associates, Inc., 2020.

\bibitem[Rebecq et~al.(2019)Rebecq, Ranftl, Koltun, and Scaramuzza]{rebecq2019high}
Henri Rebecq, Ren{\'e} Ranftl, Vladlen Koltun, and Davide Scaramuzza.
\newblock High speed and high dynamic range video with an event camera.
\newblock \emph{IEEE transactions on pattern analysis and machine intelligence}, 43\penalty0 (6):\penalty0 1964--1980, 2019.

\bibitem[Redmon(2016)]{yolo}
J Redmon.
\newblock You only look once: Unified, real-time object detection.
\newblock In \emph{Proceedings of the IEEE conference on computer vision and pattern recognition}, 2016.

\bibitem[Ross and Doll{\'a}r(2017)]{retinanet}
T-YLPG Ross and GKHP Doll{\'a}r.
\newblock Focal loss for dense object detection.
\newblock In \emph{proceedings of the IEEE conference on computer vision and pattern recognition}, pages 2980--2988, 2017.

\bibitem[Rossi et~al.(2020)Rossi, Chamberlain, Frasca, Eynard, Monti, and Bronstein]{graph2}
Emanuele Rossi, Ben Chamberlain, Fabrizio Frasca, Davide Eynard, Federico Monti, and Michael Bronstein.
\newblock Temporal graph networks for deep learning on dynamic graphs.
\newblock \emph{arXiv preprint arXiv:2006.10637}, 2020.

\bibitem[Schaefer et~al.(2022)Schaefer, Gehrig, and Scaramuzza]{aegnn}
Simon Schaefer, Daniel Gehrig, and Davide Scaramuzza.
\newblock Aegnn: Asynchronous event-based graph neural networks.
\newblock In \emph{IEEE Conference on Computer Vision and Pattern Recognition}, 2022.

\bibitem[Sekikawa et~al.(2019)Sekikawa, Hara, and Saito]{eventnet}
Yusuke Sekikawa, Kosuke Hara, and Hideo Saito.
\newblock Eventnet: Asynchronous recursive event processing.
\newblock In \emph{Proceedings of the IEEE/CVF conference on computer vision and pattern recognition}, pages 3887--3896, 2019.

\bibitem[Sironi et~al.(2018)Sironi, Brambilla, Bourdis, Lagorce, and Benosman]{sironi2018hats}
Amos Sironi, Manuele Brambilla, Nicolas Bourdis, Xavier Lagorce, and Ryad Benosman.
\newblock Hats: Histograms of averaged time surfaces for robust event-based object classification.
\newblock In \emph{Proceedings of the IEEE conference on computer vision and pattern recognition}, pages 1731--1740, 2018.

\bibitem[Su et~al.(2023)Su, Chou, Hu, Li, Mei, Zhang, and Li]{snn2}
Qiaoyi Su, Yuhong Chou, Yifan Hu, Jianing Li, Shijie Mei, Ziyang Zhang, and Guoqi Li.
\newblock Deep directly-trained spiking neural networks for object detection.
\newblock In \emph{Proceedings of the IEEE/CVF International Conference on Computer Vision}, 2023.

\bibitem[Verma et~al.(2024)Verma, Chakravarthi, Vaghela, Wei, and Yang]{verma2024etram}
Aayush~Atul Verma, Bharatesh Chakravarthi, Arpitsinh Vaghela, Hua Wei, and Yezhou Yang.
\newblock etram: Event-based traffic monitoring dataset.
\newblock In \emph{Proceedings of the IEEE/CVF Conference on Computer Vision and Pattern Recognition}, pages 22637--22646, 2024.

\bibitem[Wang et~al.(2024)Wang, Wang, Li, Qin, Jiang, Ma, and Tang]{snn4-eas-snn}
Ziming Wang, Ziling Wang, Huaning Li, Lang Qin, Runhao Jiang, De Ma, and Huajin Tang.
\newblock Eas-snn: End-to-end adaptive sampling and representation for event-based detection with recurrent spiking neural networks.
\newblock \emph{arXiv preprint arXiv:2403.12574}, 2024.

\bibitem[Yu et~al.(2024)Yu, Chen, Wang, Zhan, Shao, Liu, and Xu]{snn5}
Lixing Yu, Hanqi Chen, Ziming Wang, Shaojie Zhan, Jiankun Shao, Qingjie Liu, and Shu Xu.
\newblock Spikingvit: a multi-scale spiking vision transformer model for event-based object detection.
\newblock \emph{IEEE Transactions on Cognitive and Developmental Systems}, 2024.

\bibitem[Yuan et~al.(2024)Yuan, Zhang, Wang, Liu, Pan, and Tang]{snn1}
Mengwen Yuan, Chengjun Zhang, Ziming Wang, Huixiang Liu, Gang Pan, and Huajin Tang.
\newblock Trainable spiking-yolo for low-latency and high-performance object detection.
\newblock 2024.

\bibitem[Zheng et~al.(2023)Zheng, Liu, Lu, Hua, Pan, Zhang, Tao, and Wang]{eventvisionsurvey2}
Xu Zheng, Yexin Liu, Yunfan Lu, Tongyan Hua, Tianbo Pan, Weiming Zhang, Dacheng Tao, and Lin Wang.
\newblock Deep learning for event-based vision: A comprehensive survey and benchmarks, 2023.

\bibitem[Zheng et~al.(2021)Zheng, Wang, Ren, Liu, Ye, Hu, and Zuo]{ciou}
Zhaohui Zheng, Ping Wang, Dongwei Ren, Wei Liu, Rongguang Ye, Qinghua Hu, and Wangmeng Zuo.
\newblock Enhancing geometric factors in model learning and inference for object detection and instance segmentation.
\newblock \emph{IEEE transactions on cybernetics}, 52\penalty0 (8):\penalty0 8574--8586, 2021.

\bibitem[Zhou et~al.(2023)Zhou, Gallego, Lu, Liu, and Shen]{motionseg_graphcut}
Yi Zhou, Guillermo Gallego, Xiuyuan Lu, Siqi Liu, and Shaojie Shen.
\newblock Event-based motion segmentation with spatio-temporal graph cuts.
\newblock \emph{IEEE Transactions on Neural Networks and Learning Systems}, 34\penalty0 (8):\penalty0 4868--4880, 2023.

\bibitem[Zhu et~al.(2018)Zhu, Thakur, Özaslan, Pfrommer, Kumar, and Daniilidis]{mvsec}
Alex~Zihao Zhu, Dinesh Thakur, Tolga Özaslan, Bernd Pfrommer, Vijay Kumar, and Kostas Daniilidis.
\newblock The multivehicle stereo event camera dataset: An event camera dataset for 3d perception.
\newblock \emph{IEEE Robotics and Automation Letters}, 3\penalty0 (3):\penalty0 2032--2039, 2018.

\bibitem[Zubic et~al.(2024)Zubic, Gehrig, and Scaramuzza]{ssm}
Nikola Zubic, Mathias Gehrig, and Davide Scaramuzza.
\newblock State space models for event cameras.
\newblock In \emph{Proceedings of the IEEE/CVF Conference on Computer Vision and Pattern Recognition (CVPR)}, pages 5819--5828, 2024.

\end{thebibliography}
}

\end{document}